\documentclass[runningheads]{llncs}

\usepackage{eccv}

\usepackage{eccvabbrv}

\usepackage{graphicx}
\usepackage{booktabs}
\usepackage{multirow}
\usepackage{amsmath}
\usepackage[table]{xcolor}

\usepackage[accsupp]{axessibility}  % Improves PDF readability for those with disabilities.

\usepackage[hidelinks]{hyperref}

\usepackage{orcidlink}

\newcommand\blfootnote[1]{%
  \begingroup
  \renewcommand\thefootnote{}\footnote{#1}%
  \addtocounter{footnote}{-1}%
  \endgroup
}

\begin{document}

\title{MG-RWKV: Multi-Grained Context-Aware RWKV for Temporal Forgery Localization} 

\titlerunning{MG-RWKV}

\author{Jingchen Ni\inst{1}$^{*}$ \and
Cangjin Yu\inst{2}$^{*}$ \and
Dan Jiang\inst{1} \and
Quan Zhang\inst{1} \and
Keyu Lv\inst{1} \and
Shannan Yan\inst{1} \and
Linyue Pan\inst{1} \and
Ke Zhang\inst{2}$^{\dagger}$ \and
Chun Yuan\inst{1}$^{\dagger}$}

\authorrunning{J.~Ni, C.~Yu et al.}

\institute{$^{1}$Tsinghua University, China \quad $^{2}$Soochow University, China\\
\email{njc24@mails.tsinghua.edu.cn, zhangke@suda.edu.cn, yuanc@sz.tsinghua.edu.cn}}

\maketitle
\blfootnote{$^{*}$ Equal contribution.\quad $^{\dagger}$ Corresponding authors.}

\begin{abstract}
    \sloppy
    Driven by Artificial Intelligence-Generated Content (AIGC), the authenticity of audio-visual content is facing severe challenges. Temporal Forgery Localization (TFL) aims to precisely identify manipulated segments within untrimmed sequences. However, existing methods are limited by CNNs' local receptive fields or Transformers' quadratic complexity, while emerging linear models often struggle to balance global authentic context compression with local abrupt forgery perception. To address this, we propose MG-RWKV, a multi-granularity framework that leverages the data-dependent state evolution of RWKV to achieve efficient full-sequence processing with $\mathcal{O}(T)$ complexity. Our framework features three core innovations: (1) a \textbf{Bidirectional RWKV} architecture that captures bidirectional temporal contexts without quadratic overhead; (2) a \textbf{Multi-Granularity Mixture of Experts (MG-MoE)} that performs dynamic routing over explicit temporal receptive fields, adaptively selecting granularities based on forgery duration to significantly enhance decision interpretability; and (3) \textbf{Cross-Granularity Consistency (CGC)}, which aligns adjacent feature pyramid levels through hierarchical scale-wise pairing and spatial boundary-aware weighting, effectively reducing false positives in authentic regions. Extensive experiments on Lav-DF, TVIL, and Psynd datasets demonstrate that MG-RWKV achieves state-of-the-art performance with low computational cost.
\end{abstract}

\section{Introduction}
\label{sec:intro}

Digital content forgery detection has long stood as a pivotal focus in multimedia security\cite{el2024comprehensive,tyagi2023detailed}. Traditional forgery techniques primarily involve manipulating image data. With the rapid rise of Artificial Intelligence-Generated Content (AIGC)\cite{yu2024fake,shoaib2023deepfakes,lyu2024deepfake}, however, deepfake-driven audio-visual forgeries have emerged as the mainstream. The proliferation of such high-fidelity deceptive content raises severe societal concerns, underscoring the urgency for advanced detection technologies. Early detection approaches centered predominantly on facial forgeries\cite{patel2023deepfake,yan2023ucf,huang2023implicit}. In complex audio-visual scenarios, attackers often manipulate specific content segments through voice cloning or video tampering, producing highly deceptive material that poses significant challenges to traditional binary classification paradigms.

To address this gap, recent research redefines the task as Temporal Forgery Localization (TFL)\cite{he2021forgerynet,cai2022you}, aiming to spatially and temporally localize forged segments within untrimmed sequences. This requires the model to identify subtle manipulation traces---such as semantic replacement, emotional inconsistency, or object restoration errors---across hundreds or thousands of frames.

\begin{figure}[!t]
    \centering
    \includegraphics[width=1\linewidth]{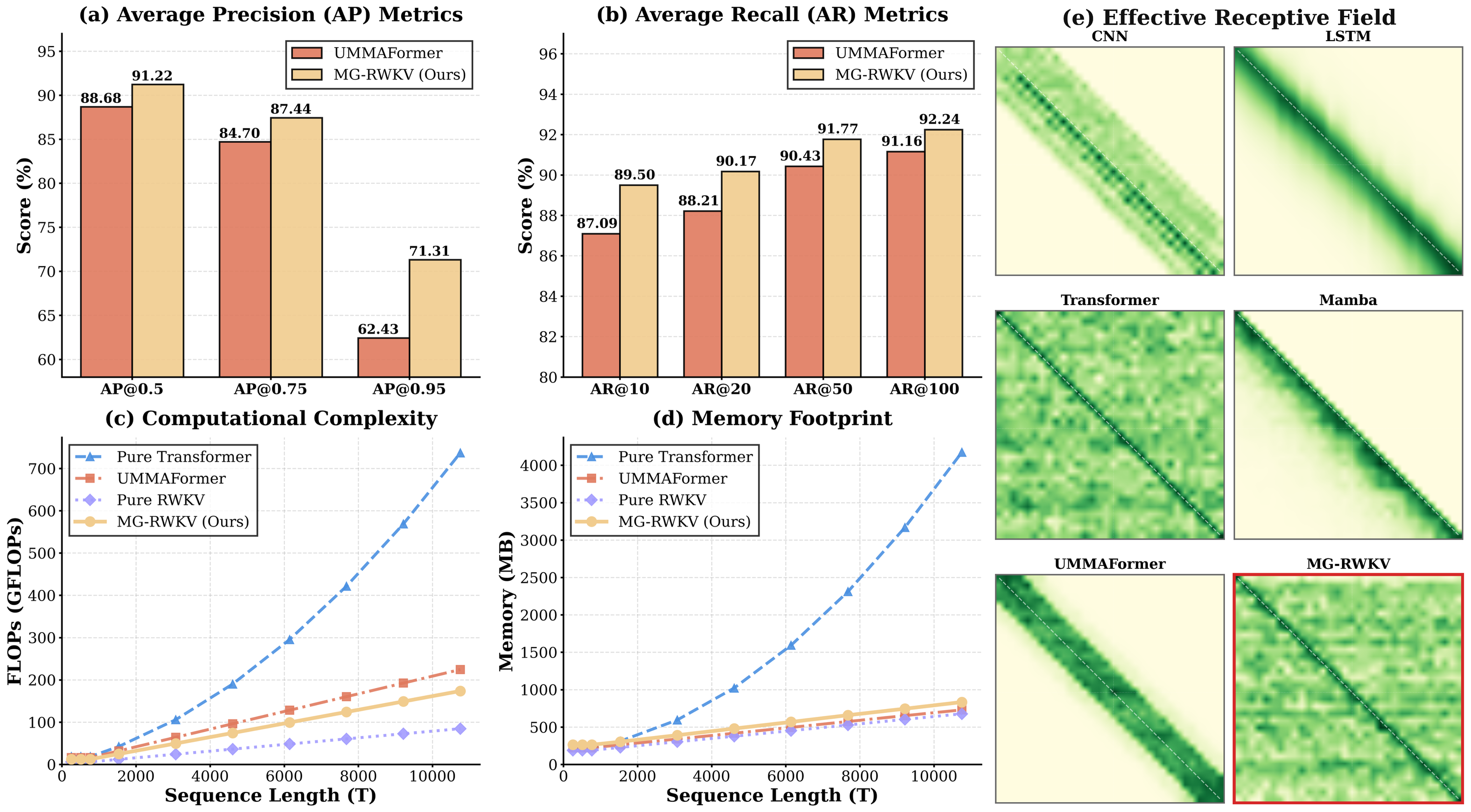}
    \caption{Performance and computational efficiency comparison on TVIL dataset. (a) Average Precision at different thresholds. (b) Average Recall at different proposal numbers. (c) Computational complexity (FLOPs) versus sequence length. (d) Memory footprint versus sequence length. (e) Effective Receptive Field (ERF) comparison across architectures—MG-RWKV exhibits dense, long-range temporal connectivity comparable to full Transformers while maintaining linear complexity. MG-RWKV achieves superior performance with linear $\mathcal{O}(T)$ scaling.}
    \label{fig:intro_comparison}
\end{figure}

However, existing TFL methods face a fundamental architectural bottleneck when modeling long-range dependencies. CNN-based approaches suffer from limited receptive fields, struggling to capture global temporal inconsistencies across time spans. Conversely, Transformer-based frameworks, while possessing global context, incur quadratic $\mathcal{O}(T^2)$ complexity via self-attention, leading to severe computational and memory bottlenecks when processing full sequences. To mitigate this, some methods\cite{zhang2023ummaformer} adopt local window attention, which inevitably sacrifices global modeling capabilities. Recently, emerging linear-complexity architectures, such as state space models (e.g., Mamba\cite{gu2024mamba}) and linear attention\cite{ma2023megamovingaverageequipped}, have shown promise. Yet, the TFL task poses a unique requirement: the model must efficiently compress the global authentic context while remaining highly sensitive to abrupt, instantaneous boundary changes caused by forgeries. Conventional linear models often struggle to achieve this optimal balance between ``global smooth compression'' and ``local abrupt perception''. We observe that the ``data-dependent decay'' and dynamic state evolution mechanisms inherent in the RWKV\cite{peng2025rwkv} architecture naturally align with this requirement, offering an ideal paradigm for TFL.

In this paper, we propose MG-RWKV, a linear-complexity framework systematically tailored for temporal forgery localization. MG-RWKV maintains linear $\mathcal{O}(T)$ scaling while enhancing the transparency and interpretability of the localization process through three synergistic modules. First, accurate forgery boundary localization depends on both ``before'' and ``after'' contexts. Building upon traditional RWKV, we design a \textbf{Bidirectional RWKV (BiDir)} architecture that simultaneously captures past and future temporal dependencies, achieving a true global receptive field without the computational burden of Transformers. As shown in \cref{fig:intro_comparison}(e), the Effective Receptive Field analysis confirms that MG-RWKV achieves dense, long-range temporal connectivity comparable to full Transformers while maintaining linear $\mathcal{O}(T)$ complexity.

Second, forgery patterns exhibit significant variations in temporal scale---ranging from instantaneous frame flickers requiring fine-grained perception to large-scale scene generations demanding a macroscopic coarse-grained view. We design a \textbf{Multi-Granularity Mixture of Experts (MG-MoE)} module. Unlike standard black-box routing, our ``experts'' are constructed from convolutional branches with different dilation rates, representing temporal receptive fields with explicit physical meanings. Through input-aware dynamic routing, the model adaptively selects the appropriate granularity based on the specific forgery duration, substantially enhancing the interpretability of the decision-making process.

Finally, to address the issue of multi-scale features producing inconsistent predictions in authentic regions—a primary source of false positives—we propose the \textbf{Cross-Granularity Consistency (CGC)} constraint. CGC achieves precise feature alignment through two core designs: structurally, it performs \textit{hierarchical scale-wise pairing} between adjacent FPN levels; and spatially, it applies \textit{boundary-aware weighting} to relax constraints at transition frames where scale-dependent differences carry genuine semantic meaning. This design effectively aligns cross-granularity representations and sharpens temporal boundary localization accuracy.

In summary, our contributions are as follows:
\begin{itemize}
    \item We propose the novel MG-RWKV framework, systematically exploring the application of a data-dependent linear recurrent architecture for the TFL task. This effectively breaks the efficiency-accuracy trade-off bottleneck of existing methods, distinguishing our approach from both Transformers and generic linear models.
    \item We design a Bidirectional RWKV architecture to capture bidirectional context and innovatively propose the MG-MoE module, which leverages dynamic routing over explicit temporal receptive fields to achieve adaptive and highly interpretable multi-scale perception.
    \item We introduce the CGC module, which significantly reduces false positives and improves boundary precision by cleanly integrating hierarchical cross-scale alignment and spatial boundary-aware weighting.
\end{itemize}

% However, limited by the local receptive field of CNNs and the quadratic complexity of Transformers, existing methods struggle to achieve efficient multi-granularity context modeling. To address this, this paper proposes MG-RWKV, a multi-granularity context-aware Receptance Weighted Key Value framework that achieves efficient temporal modeling with linear computational complexity.

% Specifically, MG-RWKV introduces three core designs: (1) a bidirectional RWKV that captures both past and future contexts while maintaining linear complexity, enabling global context awareness; (2) a Multi-Granularity Mixture of Experts (MGMoE) module that dynamically activates long- or short-term context experts according to the input, achieving adaptive multi-granularity representation learning; and (3) a progressive multi-granularity consistency mechanism with a boundary-aware weighting strategy that aligns multi-granularity representations and enhances temporal boundary localization accuracy.

% Extensive experiments on three benchmark datasets Lav-DF, Psynd, and TVIL demonstrate that MG-RWKV achieves state-of-the-art performance, significantly outperforming previous methods.

%%%%%%%%%%%%%%%%%%%%%%%%%%%%

\section{Related Work}
\label{sec:related_work}

\subsection{Image Forgery Detection}
Traditional IFD methods rely on handcrafted features such as color filter arrays, photo-response non-uniformity noise, illumination, and JPEG artifacts. Although effective in some cases, these methods struggle against advanced forgeries where manipulated regions blend seamlessly with the background. Recently, deep learning-based approaches\cite{kwon2021cat,dong2022mvss,liu2022pscc,guillaro2023trufor,zhangimdprompter,chen2025gim,ni2026fcl} have achieved remarkable progress. For instance, MVSS-Net\cite{dong2022mvss} adopts a dual-stream architecture to jointly model noise and boundary cues, PSCC-Net\cite{liu2022pscc} performs bidirectional feature aggregation, and TruFor\cite{guillaro2023trufor} fuses RGB and noise-sensitive fingerprints using a Transformer-based structure for robust trace extraction.

% Lav-DF\cite{cai2022you}, Psynd\cite{zhang2022localizing}, and TVIL\cite{zhang2023ummaformer}.
\subsection{Temporal Forgery Localization }
With the proliferation of tampered audio-visual content, accurately localizing the temporal span of forgery remains a major challenge due to data scarcity and high realism of synthetic content. To tackle this, researchers have developed benchmark datasets such as Lav-DF\cite{cai2022you} and TVIL\cite{zhang2023ummaformer} and proposed representative models. BA-TFD\cite{cai2022you} employs dual 3D CNN encoders with contrastive and boundary matching losses to capture modal desynchronization. UMMAFormer\cite{zhang2023ummaformer} introduces a Transformer-based temporal anomaly attention module and cross-attention feature pyramid network for long-range dependency modeling. More broadly, advances in cross-modal representation learning\cite{ni2025semantic} and language-guided localization\cite{wang2025iterprime} underscore the importance of aligning heterogeneous cues for precise localization.

\subsection{Temporal Action Detection}
TAD aims to identify and localize actions in untrimmed videos. Existing approaches fall into two categories: two-stage and one-stage methods. Two-stage frameworks \cite{gao2017cascaded,xu2017r} generate and classify action proposals separately, predicting action boundaries or using anchor-based strategies, but suffer from high complexity and limited end-to-end optimization. In contrast, one-stage methods \cite{lin2017single,buch2019end} jointly perform localization and classification within a unified network, achieving improved efficiency though still facing a performance gap compared with recent Transformer-based models. Beyond full supervision, weakly- and unsupervised methods \cite{Zhang_2025_CVPR,zhang2025rethinking,zhang2025eavmamba,xia2025clip} further reduce annotation cost.

\subsection{Efficient Sequence Models}
Several architectures have been proposed to replace the quadratic self-attention of Transformers with linear-complexity alternatives. Linear attention methods such as Performer \cite{performer} and Linformer \cite{wang2020linformer} approximate full attention through kernel tricks or low-rank projections, but often sacrifice modeling capacity for long-range dependencies. State Space Models (SSMs), notably S4 \cite{gu2021efficiently} and Mamba \cite{gu2024mamba}, reformulate sequence modeling as a selective state space recurrence, achieving $\mathcal{O}(T)$ complexity with competitive performance on sequence tasks. However, SSMs are designed with fixed or data-independent state transition mechanisms, which may limit their sensitivity to the subtle, locally-concentrated anomalies characteristic of forgery boundaries. RWKV \cite{peng2025rwkv} combines the efficiency of recurrent inference with data-dependent decay and in-context state modulation, providing stronger adaptive capacity for detecting temporal anomalies. Empirically, we observe that RWKV-7 outperforms Mamba on TFL (82.43 vs.\ 80.15 mAP on Lav-DF; see \cref{sec:comparison_backbone}), which we attribute to its more expressive state modulation mechanism. Importantly, MG-RWKV is not merely a substitution of Transformers with RWKV—it exploits RWKV's recurrent structure to design a dilation-based multi-granularity architecture that is not naturally supported by attention-based models.

\section{Methodology}

\subsection{Overview}

Given a feature sequence $\mathbf{X} \in \mathbb{R}^{T \times D}$ of an untrimmed video, where $T$ is the number of time steps and $D$ is the feature dimension, the goal of Temporal Forgery Localization (TFL) is to detect forged temporal segments. Following the anchor-free detection paradigm \cite{zhang2023ummaformer}, our model produces dense predictions: classification scores $\mathbf{P} \in \mathbb{R}^{T \times N_c}$ and boundary offsets $\mathbf{O} \in \mathbb{R}^{T \times 2}$ for each time position, which are then converted into segment proposals $\{(t_s^i, t_e^i, s^i)\}_{i=1}^{N}$ through post-processing.

\begin{figure*}[t]
    \centering
    \includegraphics[width=\linewidth]{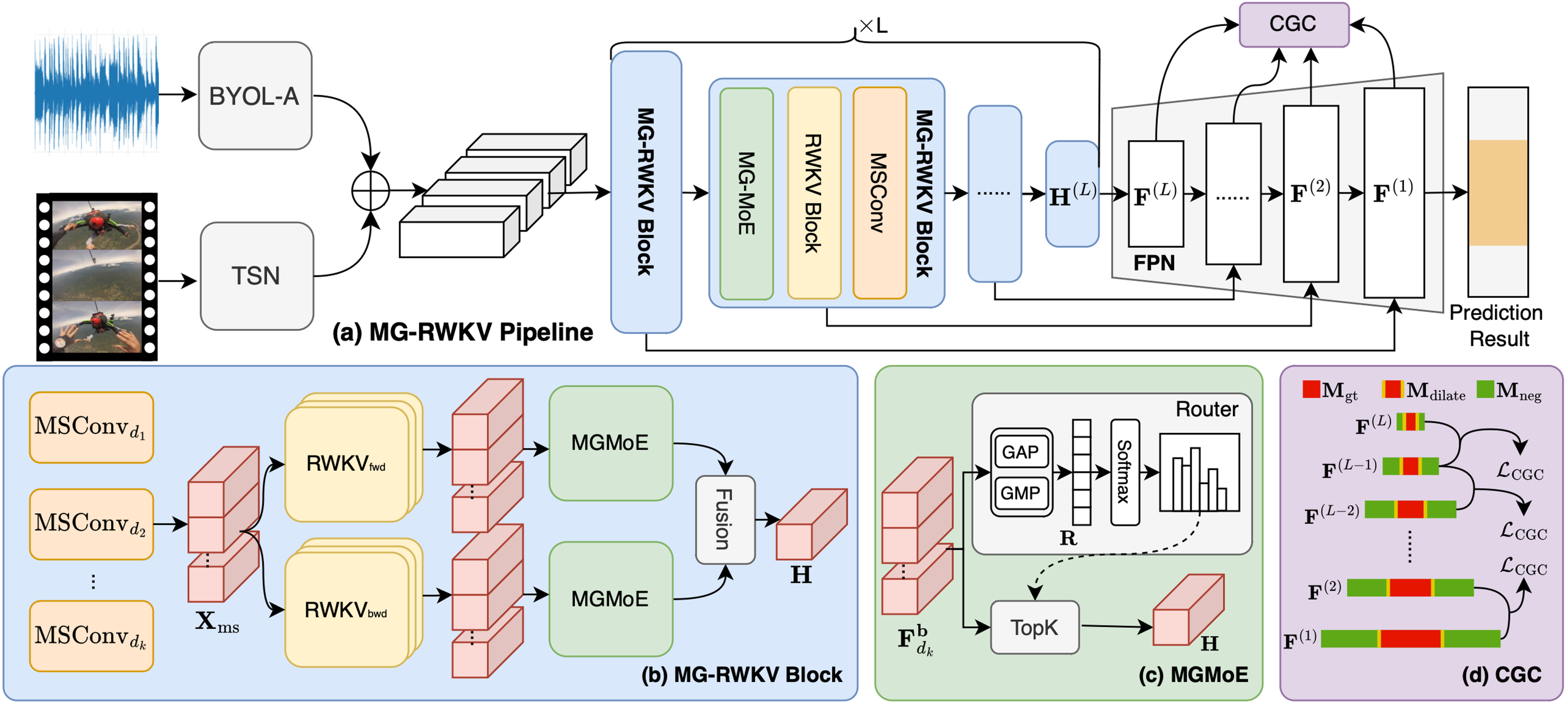}
    \caption{Overview of the proposed MG-RWKV framework. (a) Overall pipeline with BYOL-A/TSN extractors, MG-RWKV blocks, FPN, and prediction heads. (b) MG-RWKV block with multi-scale convolutions, bidirectional RWKV, and MG-MoE. (c) MG-MoE with dynamic routing via GAP/GMP and Top-K expert selection. (d) CGC for cross-granularity alignment with boundary-aware weighting, where red denotes forged regions, yellow the dilated boundary margin, and green the negative regions.}
    \label{fig:framework}
\end{figure*}

As illustrated in \cref{fig:framework}, MG-RWKV first extracts visual and audio features using pre-trained TSN \cite{wang2016temporal} and BYOL-A \cite{niizumi2021byol}, which are fused and projected to form the input sequence $\mathbf{X}$. The sequence is then processed by $L$ stacked MG-RWKV blocks to produce hierarchical multi-scale features $\{\mathbf{H}^{(l)}\}_{l=1}^L$, where each block applies dilated multi-scale convolution and bidirectional RWKV with MG-MoE routing. A top-down Feature Pyramid Network (FPN) further refines and fuses these features into $\{\mathbf{F}^{(l)}\}_{l=1}^L$, upon which classification and regression heads output dense forgery scores and boundary offsets. Finally, we apply Soft-NMS \cite{bodla2017soft} to convert dense predictions into the top-100 segment proposals.

The framework incorporates three core innovations. The \textbf{Bidirectional RWKV Architecture} replaces quadratic self-attention with a linear-complexity recurrent mechanism, while the bidirectional scan provides global temporal context essential for boundary localization. The \textbf{Multi-Granularity Mixture of Experts (MG-MoE)} treats BiRWKV branches with structurally distinct dilation rates as interpretable experts, enabling position-adaptive granularity selection through dynamic routing. The \textbf{Cross-Granularity Consistency (CGC)} enforces cross-scale feature agreement in authentic regions, resolving the cross-granularity contradictions that multi-scale modeling inherently introduces. These three components form a closed-loop design: BiDir establishes the global context upon which MG-MoE performs adaptive multi-scale routing, while CGC eliminates the inter-scale inconsistencies that such routing would otherwise introduce.

\subsection{RWKV-7 for Temporal Modeling}

Accurate boundary localization in TFL requires both past and future context: a unidirectional model cannot exploit post-boundary information when predicting segment starts, leading to systematically imprecise boundaries. Standard bidirectional Transformers address this via full self-attention, but incur $\mathcal{O}(T^2 \cdot d)$ complexity—for Lav-DF videos with $T \approx 1500$, this yields ${\sim}2.25 \times 10^6$ pairwise computations per layer and becomes prohibitive at scale.

\noindent \textbf{RWKV-7 Architecture.} We adopt RWKV-7 \cite{peng2025rwkv}, a linear-complexity recurrent model with data-dependent decay and in-context state modulation. Given input $\mathbf{x}_t \in \mathbb{R}^d$, RWKV-7 applies token shift mixing, then computes receptance $\mathbf{r}_t$, key $\mathbf{k}_t$, value $\mathbf{v}_t$, and gate $\mathbf{g}_t$ via linear projections. The key innovation generates adaptive parameters through input-dependent quadratic functions:

\begin{equation}
\label{eq:rwkv_params}
\begin{aligned}
d_t &= \mathbf{w}_0 + \mathbf{w}_1 \odot \mathbf{x}_t' + \mathbf{w}_2 \odot (\mathbf{x}_t')^2 \\
\mathbf{a}_t &= \mathbf{a}_0 + \mathbf{a}_1 \odot \mathbf{x}_t'' + \mathbf{a}_2 \odot (\mathbf{x}_t'')^2
\end{aligned}
\end{equation}

\noindent where $\mathbf{x}_t', \mathbf{x}_t''$ are token-shifted inputs, $\mathbf{w}_i, \mathbf{a}_i$ are learnable parameters, and $d_t$ controls the per-step decay strength. The recurrent state evolves as:

\begin{equation}
\label{eq:rwkv_state}
\mathbf{s}_t = e^{-e^{d_t}} \odot \mathbf{s}_{t-1} + \mathbf{k}_t \odot \mathbf{v}_t + \mathbf{a}_t \odot \mathbf{s}_{t-1}, \quad \mathbf{o}_t = \mathbf{r}_t \odot \mathbf{s}_t \cdot \sigma(\mathbf{g}_t)
\end{equation}

\noindent where $e^{-e^{d_t}}$ provides numerically stable exponential decay and $\mathbf{a}_t$ enables in-context state modulation. Since $\mathbf{o}_t$ depends only on $\mathbf{x}_t$ and $\mathbf{s}_{t-1}$, the overall complexity is $\mathcal{O}(T \cdot d^2)$—linear in sequence length. Each RWKV-7 block interleaves this Time Mix module with a ReLU$^2$ MLP under pre-normalization and residual connections.

\noindent \textbf{Bidirectional Extension.} To simultaneously capture past and future context while maintaining linear complexity, we extend RWKV-7 bidirectionally by applying forward and backward scans with independent parameter sets $\boldsymbol{\theta}^{\text{fwd}}$ and $\boldsymbol{\theta}^{\text{bwd}}$. The resulting forward features $\{\mathbf{F}_k^{\text{fwd}}\}$ and backward features $\{\mathbf{F}_k^{\text{bwd}}\}$ are subsequently fused through the Multi-Granularity Mixture of Experts mechanism described in the following section.

% [Figure 2: RWKV-7 block architecture showing LayerNorm, Time Mix, and ReLU² MLP components]

\subsection{Multi-Granularity Mixture of Experts}

Video forgeries span a wide range of temporal scales: frame-level flickers demand fine-grained local analysis, while long-duration synthesis requires coarse-grained global context. Rather than treating this as an open-ended search over arbitrary resolutions, we observe that forgery temporal scales form a structured spectrum—analogous to how spatial object scales cluster around characteristic sizes in detection tasks. MG-MoE operationalizes this observation by defining each expert as a BiRWKV branch with a structurally distinct dilation rate, so the temporal receptive field of every expert is an explicit, interpretable quantity rather than an emergent property of unconstrained learned weights. The appropriate granularity varies by position, motivating a data-driven routing mechanism that selects among experts conditioned on local temporal content.

\noindent \textbf{Scale-Structured Expert Bank.} The forgery scale spectrum is discretized into $K$ representative levels through dilation rates $\mathcal{D} = \{d_1, d_2, \ldots, d_K\}$. Input features $\mathbf{X} \in \mathbb{R}^{T \times C}$ are first enriched with multi-scale local context via a gated depthwise-dilated convolution:

\begin{equation}
\label{eq:msconv}
\mathbf{X}_{\text{ms}} = \mathbf{X} + \gamma \cdot \text{MSConv}_{\mathcal{D}}(\mathbf{X})
\end{equation}

\noindent where $\gamma$ is a learnable gate controlling injection strength, and $\text{MSConv}_{\mathcal{D}}$ fuses local multi-scale information across all rates in $\mathcal{D}$ before the expert split. Each branch $k$ then processes $\mathbf{X}_{\text{ms}}$ bidirectionally at dilation $d_k$, yielding an effective temporal receptive field of $(w{-}1) \times d_k + 1$ frames, where $w$ is the kernel size:

\begin{equation}
\label{eq:birwkv}
\begin{aligned}
\mathbf{F}_k^{\text{fwd}} &= \text{RWKV}_{d_k}^{\text{fwd}}(\mathbf{X}_{\text{ms}}) \\
\mathbf{F}_k^{\text{bwd}} &= \text{flip}\Big(\text{RWKV}_{d_k}^{\text{bwd}}\big(\text{flip}(\mathbf{X}_{\text{ms}})\big)\Big)
\end{aligned}
\end{equation}

\noindent This yields $2K$ expert representations $\{\mathbf{F}_k^{\text{fwd}}\}_{k=1}^K$ and $\{\mathbf{F}_k^{\text{bwd}}\}_{k=1}^K$, where each expert encodes forgery evidence at a distinct temporal resolution.

\noindent \textbf{Position-Adaptive Scale Selection.} The routing objective is to estimate the scale preference at each temporal position from the current expert activations. Because forward and backward scans accumulate different contextual histories, they may form different scale preferences at the same position; we therefore compute independent routing weights per direction. To capture both the overall response magnitude and the presence of discriminative anomaly spikes, we represent each expert's activation by the channel-wise mean and maximum responses:

\begin{equation}
\label{eq:pooling}
\begin{aligned}
\mathbf{R}_{\text{mean}} &= \Big[\text{Mean}_C(\mathbf{F}_k^{\text{fwd}}), \text{Mean}_C(\mathbf{F}_k^{\text{bwd}})\Big]_{k=1}^K \in \mathbb{R}^{T \times 2K} \\
\mathbf{R}_{\text{max}} &= \Big[\text{Max}_C(\mathbf{F}_k^{\text{fwd}}), \text{Max}_C(\mathbf{F}_k^{\text{bwd}})\Big]_{k=1}^K \in \mathbb{R}^{T \times 2K}
\end{aligned}
\end{equation}

\noindent Mean pooling summarizes the broadband activation energy while max pooling preserves the most salient anomaly signals. Together they form a compact representation that captures both average response level and peak discriminative evidence—providing the router with complementary views for reliable scale selection. A lightweight 1D convolution with temperature-scaled softmax translates this into time-varying routing weights:

\begin{equation}
\label{eq:router_weights}
\mathbf{W}^{\mathbf{b}} = \text{softmax}\!\left(\frac{\text{Conv1D}([\mathbf{R}_{\text{mean}} \oplus \mathbf{R}_{\text{max}}])^{\mathbf{b}}}{\tau}\right) \in \mathbb{R}^{T \times K}
\end{equation}

\noindent where $\mathbf{b} \in \{\text{fwd}, \text{bwd}\}$ and temperature $\tau$ governs the sharpness of scale selection. To prevent expert collapse—wherein a dense soft mixture over all scales would incentivize experts to converge toward similar average representations—we apply sparse Top-$K_{\text{top}}$ gating, enforcing that each position activates only a subset of experts:

\begin{equation}
\label{eq:topk}
\text{TopK}(\mathbf{W}, K_{\text{top}}) = \frac{\mathbf{W} \odot \mathbb{1}_{\text{top-}K_{\text{top}}}}{\sum_{k} \mathbf{W}_k \odot \mathbb{1}_{\text{top-}K_{\text{top}}}}
\end{equation}

\noindent where $\mathbb{1}_{\text{top-}K_{\text{top}}}$ retains only the $K_{\text{top}}$ largest weights per position, enforcing specialization and maintaining representational diversity across experts. The weighted aggregation then yields direction-specific fused representations:

\begin{equation}
\label{eq:weighted_sum}
\begin{aligned}
\tilde{\mathbf{W}}^{\mathbf{b}} &= \text{TopK}(\mathbf{W}^{\mathbf{b}}, K_{\text{top}}), \\
\mathbf{H}^{\mathbf{b}} &= \sum_{k=1}^K \tilde{\mathbf{W}}_k^{\mathbf{b}} \odot \mathbf{F}_k^{\mathbf{b}}, \quad \mathbf{b} \in \{\text{fwd}, \text{bwd}\}
\end{aligned}
\end{equation}

\noindent The two direction-specific outputs $\mathbf{H}^{\text{fwd}}$ and $\mathbf{H}^{\text{bwd}}$ are then fused via linear projection:

\begin{equation}
\label{eq:fusion}
\mathbf{H} = W^{\text{fusion}}[\mathbf{H}^{\text{fwd}} \oplus \mathbf{H}^{\text{bwd}}]
\end{equation}

\noindent where $W^{\text{fusion}} \in \mathbb{R}^{C \times 2C}$ projects the concatenated bidirectional representations back to dimension $C$. Setting $K_{\text{top}}{=}2$ permits adjacent granularities to be jointly activated at boundary positions, enabling smooth scale transitions that a hard top-1 selection would suppress.

% [Figure 3: MGMoE module structure and routing weight visualization in different scenarios]

\subsection{Cross-Granularity Consistency}

While MG-MoE captures multi-scale forgery patterns effectively, parallel branches with heterogeneous receptive fields can produce inconsistent predictions in authentic regions, elevating false positives. CGC addresses this by enforcing cosine similarity between adjacent FPN scale features exclusively in authentic regions, preserving scale-specific discriminative capacity in forged regions while suppressing cross-scale contradictions elsewhere.

Given backbone outputs $\{\mathbf{H}^{(l)}\}_{l=1}^L$, the FPN performs top-down fusion:

\begin{equation}
\label{eq:fpn}
\begin{aligned}
\mathbf{F}^{(l)} &= \text{Conv}(\mathbf{H}^{(l)} + \text{Upsample}(\mathbf{F}^{(l+1)})), \quad l = L-1, \ldots, 1 \\
\mathbf{F}^{(L)} &= \text{Conv}(\mathbf{H}^{(L)})
\end{aligned}
\end{equation}

The authentic region mask is constructed by dilating the ground-truth forgery mask $\mathbf{M}_{\text{gt}}$ by radius $r$: $\mathbf{M}_{\text{dilate}} = \text{MaxPool1D}(\mathbf{M}_{\text{gt}}, 2r{+}1)$, then taking the complement within valid positions: $\mathbf{M}_{\text{neg}} = \mathbf{M}_{\text{valid}} \land \neg \mathbf{M}_{\text{dilate}}$. Boundary-aware weights $\mathbf{W}_b$ further reduce the constraint strength to 0.5 within $r_b$ frames of segment boundaries and maintain 1.0 elsewhere, acknowledging that near-boundary frames exhibit genuine scale-dependent transition behaviors.

For adjacent FPN scale pairs $\mathcal{P} = \{(l, l{+}1)\}_{l=1}^{L-1}$, the consistency loss is:

\begin{equation}
\label{eq:pmgc_loss}
\mathcal{L}_{\text{CGC}} = \frac{1}{|\mathcal{P}|} \sum_{(i,j) \in \mathcal{P}} \frac{\sum_{t} \mathbf{M}_{\text{neg}}(t) \cdot \mathbf{W}_b(t) \cdot d_{\text{cos}}(\mathbf{F}^{(i)}_t, \mathbf{F}^{(j)}_t)}{\sum_t \mathbf{M}_{\text{neg}}(t)}
\end{equation}

\noindent where $d_{\text{cos}}(\mathbf{a}, \mathbf{b}) = 1 - \frac{\mathbf{a} \cdot \mathbf{b}}{\|\mathbf{a}\| \|\mathbf{b}\|}$. Applying this constraint from the very first epoch risks collapsing multi-scale diversity before features have developed meaningful representations. We therefore introduce a progressive warmup schedule:

\begin{equation}
\label{eq:warmup}
\lambda_{\text{CGC}}(e) = \begin{cases}
\lambda_0 \cdot e / E_w, & e \leq E_w \\
\lambda_0, & e > E_w
\end{cases}
\end{equation}

\noindent where $e$ is the current epoch, $E_w$ is the warmup duration, and $\lambda_0$ is the target weight. Together, these three design dimensions of CGC reinforce each other: hierarchical scale-wise pairing propagates consistency locally between adjacent levels rather than collapsing all scales simultaneously; boundary-aware weighting relaxes constraints at transition frames where scale-dependent differences carry semantic meaning; and epoch-wise warmup defers enforcement until each scale has developed its own discriminative representation.

% [Figure 4: Illustration of PMGC mechanism and warmup effect comparison]

\subsection{Training Objective}

The total training loss combines classification, regression, reconstruction, and consistency objectives:

\begin{equation}
\label{eq:total_loss}
\mathcal{L}_{\text{total}} = \mathcal{L}_{\text{cls}} + \lambda_{\text{reg}} \mathcal{L}_{\text{reg}} + \mathcal{L}_{\text{reco}} + \lambda_{\text{CGC}}(e) \mathcal{L}_{\text{CGC}}
\end{equation}

\noindent where Focal Loss $\mathcal{L}_{\text{cls}}$ handles class imbalance, DIoU Loss $\mathcal{L}_{\text{reg}}$ optimizes boundary localization, $\mathcal{L}_{\text{reco}}$ is an auxiliary reconstruction objective, and $\mathcal{L}_{\text{CGC}}$ enforces multi-scale consistency under the warmup schedule of \cref{eq:warmup}.

\begin{table*}[!t]
\centering
\caption{Comparison with state-of-the-art methods on Lav-DF, TVIL, and Psynd datasets. AP and AR denote Average Precision and Average Recall at the specified tIoU thresholds. The best result per column is \textbf{bold}; MG-RWKV rows are highlighted in gray.}
\label{tab:my-table_01}
\resizebox{1.00\textwidth}{!}{%
\begin{tabular}{cccccccccc}
\hline
\multicolumn{1}{c|}{Dataset} &
  \multicolumn{1}{c|}{Methods} &
  \multicolumn{1}{c|}{Feature} &
  AP@0.5 &
  AP@0.75 &
  AP@0.95 &
  AR@10 &
  AR@20 &
  AR@50 &
  AR@100 \\ \hline
\multicolumn{1}{c|}{\multirow{15}{*}{Lav-DF}} &
\multicolumn{1}{c|}{MDS \cite{chugh2020not}} &
  \multicolumn{1}{c|}{Visual} &
  12.78 &
  1.62 &
  0.00 &
  37.88 &
  36.71 &
  34.39 &
  32.15 \\
\multicolumn{1}{c|}{} &
\multicolumn{1}{c|}{AGT \cite{nawhal2021activity}} &
  \multicolumn{1}{c|}{Visual} &
  17.85 &
  9.42 &
  0.11 &
  43.15 &
  34.23 &
  24.59 &
  16.71 \\
\multicolumn{1}{c|}{} &
\multicolumn{1}{c|}{BMN \cite{lin2019bmn}} &
  \multicolumn{1}{c|}{Visual} &
  24.01 &
  7.61 &
  0.07 &
  53.26 &
  41.24 &
  31.60 &
  26.93 \\
\multicolumn{1}{c|}{} &
\multicolumn{1}{c|}{BMN (I3D) \cite{lin2019bmn}} &
  \multicolumn{1}{c|}{Visual} &
  10.56 &
  1.66 &
  0.00 &
  48.49 &
  44.39 &
  37.13 &
  31.55 \\
\multicolumn{1}{c|}{} &
\multicolumn{1}{c|}{AVFusion \cite{bagchi2021hear}} &
  \multicolumn{1}{c|}{Visual+Audio} &
  65.38 &
  23.89 &
  0.11 &
  62.98 &
  59.26 &
  54.80 &
  52.11 \\
\multicolumn{1}{c|}{} &
\multicolumn{1}{c|}{BA-TFD \cite{cai2022you}} &
  \multicolumn{1}{c|}{Visual} &
  58.55 &
  28.60 &
  0.16 &
  62.49 &
  58.77 &
  53.86 &
  50.29 \\
\multicolumn{1}{c|}{} &
\multicolumn{1}{c|}{BA-TFD \cite{cai2022you}} &
  \multicolumn{1}{c|}{Visual+Audio} &
  76.90 &
  38.50 &
  0.25 &
  66.90 &
  64.08 &
  60.77 &
  58.42 \\
\multicolumn{1}{c|}{} &
\multicolumn{1}{c|}{ActionFormer \cite{zhang2022actionformer}} &
  \multicolumn{1}{c|}{Visual} &
  95.34 &
  90.20 &
  23.73 &
  88.41 &
  89.63 &
  90.33 &
  90.41 \\
\multicolumn{1}{c|}{} &
\multicolumn{1}{c|}{UMMAFormer \cite{zhang2023ummaformer}} &
  \multicolumn{1}{c|}{Visual} &
  97.30 &
  92.96 &
  25.68 &
  90.19 &
  90.85 &
  91.14 &
  91.18 \\
\multicolumn{1}{c|}{} &
\multicolumn{1}{c|}{UMMAFormer \cite{zhang2023ummaformer}} &
  \multicolumn{1}{c|}{Visual+Audio} &
  98.83 &
  \textbf{95.54} &
  37.61 &
  \textbf{92.10} &
  92.42 &
  92.47 &
  92.48 \\
%\multicolumn{1}{c|}{} &
%\multicolumn{1}{c|}{DiMoDif \cite{dimodif2024}} &
%  \multicolumn{1}{c|}{Visual+Audio} &
%  95.50 &
%  87.90 &
%  20.60 &
%  91.40 &
%  92.70 &
%  93.70 &
%  94.20 \\
\multicolumn{1}{c|}{} &
\multicolumn{1}{c|}{TriDet \cite{shi2023tridet}} &
  \multicolumn{1}{c|}{Visual+Audio} &
  96.29 &
  86.84 &
  23.64 &
  88.69 &
  89.71 &
  90.39 &
  91.00 \\
\multicolumn{1}{c|}{} &
\multicolumn{1}{c|}{MFMS \cite{zhang2024mfms}} &
  \multicolumn{1}{c|}{Visual+Audio} &
  98.47 &
  94.15 &
  27.80 &
  90.02 &
  90.46 &
  90.65 &
  90.69 \\
\multicolumn{1}{c|}{} &
\multicolumn{1}{c|}{ICS-AV \cite{anshul2025intra}} &
  \multicolumn{1}{c|}{Visual+Audio} &
  87.40 &
  66.80 &
  5.72 &
  —— &
  —— &
  —— &
  —— \\
\multicolumn{1}{c|}{} &
\multicolumn{1}{c|}{\cellcolor[HTML]{DEE0E3}\textbf{MG-RWKV}} &
  \multicolumn{1}{c|}{\cellcolor[HTML]{DEE0E3}\textbf{Visual}} &
  \cellcolor[HTML]{DEE0E3}96.73 &
  \cellcolor[HTML]{DEE0E3}92.36 &
  \cellcolor[HTML]{DEE0E3}26.60 &
  \cellcolor[HTML]{DEE0E3}90.14 &
  \cellcolor[HTML]{DEE0E3}91.18 &
  \cellcolor[HTML]{DEE0E3}92.17 &
  \cellcolor[HTML]{DEE0E3}92.17 \\
\multicolumn{1}{c|}{} &
\multicolumn{1}{c|}{\cellcolor[HTML]{DEE0E3}\textbf{MG-RWKV}} &
  \multicolumn{1}{c|}{\cellcolor[HTML]{DEE0E3}\textbf{Visual+Audio}} &
  \cellcolor[HTML]{DEE0E3}\textbf{98.92} &
  \cellcolor[HTML]{DEE0E3}94.81 &
  \cellcolor[HTML]{DEE0E3}\textbf{38.47} &
  \cellcolor[HTML]{DEE0E3}91.64 &
  \cellcolor[HTML]{DEE0E3}\textbf{92.45} &
  \cellcolor[HTML]{DEE0E3}\textbf{93.41} &
  \cellcolor[HTML]{DEE0E3}\textbf{93.41} \\ \hline
\multicolumn{1}{c|}{\multirow{5}{*}{TVIL}} &
\multicolumn{1}{c|}{TAGS \cite{nag2022proposal}} &
  \multicolumn{1}{c|}{Visual} &
  18.40 &
  12.68 &
  0.09 &
  24.41 &
  25.05 &
  25.56 &
  25.56 \\
\multicolumn{1}{c|}{} &
\multicolumn{1}{c|}{DCAN \cite{chen2022dcan}} &
  \multicolumn{1}{c|}{Visual} &
  82.75 &
  75.00 &
  3.22 &
  64.73 &
  66.02 &
  68.82 &
  69.97 \\
\multicolumn{1}{c|}{} &
\multicolumn{1}{c|}{ActionFormer \cite{zhang2022actionformer}} &
  \multicolumn{1}{c|}{Visual} &
  86.27 &
  83.03 &
  28.17 &
  84.82 &
  85.77 &
  88.10 &
  88.49 \\
\multicolumn{1}{c|}{} &
\multicolumn{1}{c|}{UMMAFormer\cite{zhang2023ummaformer}} &
  \multicolumn{1}{c|}{Visual} &
  88.68 &
  84.70 &
  62.43 &
  87.09 &
  88.21 &
  90.43 &
  91.16 \\
\multicolumn{1}{c|}{} &
\multicolumn{1}{c|}{\cellcolor[HTML]{DEE0E3}\textbf{MG-RWKV}} &
  \multicolumn{1}{c|}{\cellcolor[HTML]{DEE0E3}\textbf{Visual}} &
  \cellcolor[HTML]{DEE0E3}\textbf{91.22} &
  \cellcolor[HTML]{DEE0E3}\textbf{87.44} &
  \cellcolor[HTML]{DEE0E3}\textbf{71.31} &
  \cellcolor[HTML]{DEE0E3}\textbf{89.50} &
  \cellcolor[HTML]{DEE0E3}\textbf{90.17} &
  \cellcolor[HTML]{DEE0E3}\textbf{91.77} &
  \cellcolor[HTML]{DEE0E3}\textbf{92.24} \\ \hline
\multicolumn{1}{c|}{\multirow{2}{*}{Psynd}} &
\multicolumn{1}{c|}{UMMAFormer\cite{zhang2023ummaformer}} &
  \multicolumn{1}{c|}{Audio} &
  100.00 &
  \textbf{100.00} &
  79.87 &
  97.60 &
  97.60 &
  97.60 &
  97.60 \\
\multicolumn{1}{c|}{} &
\multicolumn{1}{c|}{\cellcolor[HTML]{DEE0E3}\textbf{MG-RWKV}} &
  \multicolumn{1}{c|}{\cellcolor[HTML]{DEE0E3}\textbf{Audio}} &
  \cellcolor[HTML]{DEE0E3}\textbf{100.00} &
  \cellcolor[HTML]{DEE0E3}98.38 &
  \cellcolor[HTML]{DEE0E3}\textbf{90.09} &
  \cellcolor[HTML]{DEE0E3}\textbf{98.61} &
  \cellcolor[HTML]{DEE0E3}\textbf{98.61} &
  \cellcolor[HTML]{DEE0E3}\textbf{98.61} &
  \cellcolor[HTML]{DEE0E3}\textbf{98.61}\\
  \hline
\end{tabular}
}
\end{table*}

\section{Experiment}
\label{experiment}

\subsection{Experimental Setup}

\noindent \textbf{Datasets.} We conduct experiments on three benchmark datasets covering diverse forgery scenarios. \textbf{Lav-DF} \cite{cai2022you} is a multi-modal audio-visual dataset built upon VoxCeleb2 \cite{chung2018voxceleb2}, featuring content-driven deepfake forgeries. \textbf{TVIL} \cite{zhang2023ummaformer} is a video-only dataset derived from YouTubeVOS 2018 \cite{xu2018youtube}, containing forgeries generated via video inpainting. \textbf{Psynd} \cite{zhang2022localizing} is an audio-only dataset based on LibriTTS \cite{zen2019libritts}, featuring voice cloning forgeries.

\noindent \textbf{Evaluation Metrics.} Following prior works~\cite{cai2022you,he2021forgerynet}, we adopt Average Precision (AP) and Average Recall (AR) as the main metrics, with the tIoU thresholds set to $\{0.5, 0.75, 0.95\}$ for AP and the Average Number of proposals (AN) set to $\{10, 20, 50, 100\}$ for AR. For Psynd, we additionally report tIoU-based results following its official protocol.

\noindent \textbf{Implementation Details.} Visual and audio features are extracted using pre-trained TSN\cite{wang2016temporal} and BYOL-A\cite{niizumi2021byol}. MG-RWKV adopts embedding dimension $C=256$, pyramid blocks [2,2,5], dilation rates $\{1,2,4\}$, and convolution kernel size $w=3$. MG-MoE uses temperature $\tau=0.9$ and Top-K $K_{\text{top}}=2$; CGC employs ignore radius $r=8$, boundary radius $r_b=6$, and warmup epochs $E_{\text{warmup}}=5$. Training uses AdamW\cite{loshchilov2017fixing} with initial learning rate $\eta_0=10^{-4}$ and cosine annealing for 45 epochs on Lav-DF and TVIL, and 30 epochs on Psynd. Loss weights are $\lambda_{\text{reg}}=2.0$ and $\lambda_0=0.01$. Data augmentation includes random cropping, label smoothing, and drop path. During inference, Soft-NMS\cite{bodla2017soft} retains the top-100 proposals. All experiments are conducted on NVIDIA RTX 3090 GPUs.

% The implementation follows existing research paradigms, with specific settings as below:

% Feature Extraction: Visual features are extracted using a pre-trained two-stream TSN network [50], with a feature dimension of 4096. Audio features are extracted using a dataset-adapted model: for the Psynd dataset, a pre-trained BYOL-A model [51] (feature dimension: 2048) is used. The temporal dimension of all features is interpolated to 768 to unify the input scale.

% Hyperparameter Settings: Differentiated configurations are adopted to adapt to the characteristics of different datasets. For the learning rate: 1e-2 is used for the Lav-DF and Psynd datasets, while 2e-2 is used for the TVIL dataset. Weight decay is uniformly set to 1e-5 across all datasets.

\begin{table*}[!t]
\centering
\caption{Progressive component ablation on Lav-DF, TVIL, and Psynd datasets. Baseline ($\times\times\times$) denotes unidirectional RWKV-7 with FPN but without BiDir, MG-MoE, or CGC. $\checkmark$/$\times$ indicates whether each component is included.}
\label{tab:ablation_combined}
\resizebox{0.95\textwidth}{!}{%
\begin{tabular}{cccccccccccc}
\hline
\multicolumn{1}{c|}{Dataset} & \multicolumn{1}{c}{BiDir} & \multicolumn{1}{c}{MG-MoE} & \multicolumn{1}{c|}{CGC} & 
mAP & AP@0.5 & AP@0.75 & AP@0.95 & AR@10 & AR@20 & AR@50 & AR@100 \\ \hline
\multicolumn{1}{c|}{\multirow{4}{*}{Lav-DF}} & \multicolumn{1}{c}{$\times$} & \multicolumn{1}{c}{$\times$} & \multicolumn{1}{c|}{$\times$} & 
82.43 & 97.88 & 91.11 & 27.38 & 88.20 & 89.19 & 90.58 & 91.64 \\
\multicolumn{1}{c|}{} & \multicolumn{1}{c}{\checkmark} & \multicolumn{1}{c}{$\times$} & \multicolumn{1}{c|}{$\times$} & 
85.99 & 98.47 & 93.55 & 36.25 & 90.86 & 91.73 & 93.01 & 93.25 \\
\multicolumn{1}{c|}{} & \multicolumn{1}{c}{\checkmark} & \multicolumn{1}{c}{\checkmark} & \multicolumn{1}{c|}{$\times$} & 
86.94 & 98.89 & 94.50 & 37.31 & 91.44 & 92.26 & 93.27 & 93.31 \\
\multicolumn{1}{c|}{} & \multicolumn{1}{c}{\checkmark} & \multicolumn{1}{c}{\checkmark} & \multicolumn{1}{c|}{\checkmark} & 
\textbf{87.29} & \textbf{98.92} & \textbf{94.81} & \textbf{38.47} & \textbf{91.64} & \textbf{92.45} & \textbf{93.41} & \textbf{93.41} \\ \hline
\multicolumn{1}{c|}{\multirow{4}{*}{TVIL}} & \multicolumn{1}{c}{$\times$} & \multicolumn{1}{c}{$\times$} & \multicolumn{1}{c|}{$\times$} & 
83.32 & 89.20 & 85.55 & 63.37 & 87.44 & 88.77 & 90.13 & 90.86 \\
\multicolumn{1}{c|}{} & \multicolumn{1}{c}{\checkmark} & \multicolumn{1}{c}{$\times$} & \multicolumn{1}{c|}{$\times$} & 
83.08 & 89.11 & 85.95 & 66.58 & 87.20 & 88.58 & 90.50 & 91.01 \\
\multicolumn{1}{c|}{} & \multicolumn{1}{c}{\checkmark} & \multicolumn{1}{c}{\checkmark} & \multicolumn{1}{c|}{$\times$} & 
84.35 & 90.43 & 87.42 & 65.87 & 88.38 & 88.99 & 90.65 & 91.57 \\
\multicolumn{1}{c|}{} & \multicolumn{1}{c}{\checkmark} & \multicolumn{1}{c}{\checkmark} & \multicolumn{1}{c|}{\checkmark} & 
\textbf{85.91} & \textbf{91.22} & \textbf{87.44} & \textbf{71.31} & \textbf{89.50} & \textbf{90.17} & \textbf{91.77} & \textbf{92.24} \\ \hline
\multicolumn{1}{c|}{\multirow{4}{*}{Psynd}} & \multicolumn{1}{c}{$\times$} & \multicolumn{1}{c}{$\times$} & \multicolumn{1}{c|}{$\times$} & 
92.49 & 100.00 & 95.68 & 68.69 & 95.57 & 95.57 & 95.57 & 95.57 \\
\multicolumn{1}{c|}{} & \multicolumn{1}{c}{\checkmark} & \multicolumn{1}{c}{$\times$} & \multicolumn{1}{c|}{$\times$} & 
96.21 & 100.00 & 97.89 & 75.84 & 97.47 & 97.47 & 97.47 & 97.47 \\
\multicolumn{1}{c|}{} & \multicolumn{1}{c}{\checkmark} & \multicolumn{1}{c}{\checkmark} & \multicolumn{1}{c|}{$\times$} & 
97.67 & 100.00 & 98.20 & 87.18 & 98.35 & 98.35 & 98.35 & 98.35 \\
\multicolumn{1}{c|}{} & \multicolumn{1}{c}{\checkmark} & \multicolumn{1}{c}{\checkmark} & \multicolumn{1}{c|}{\checkmark} & 
\textbf{98.23} & \textbf{100.00} & \textbf{98.38} & \textbf{90.09} & \textbf{98.61} & \textbf{98.61} & \textbf{98.61} & \textbf{98.61} \\ \hline
\end{tabular}
}
\end{table*}

\begin{figure}[!t]
\centering
\includegraphics[width=1\linewidth]{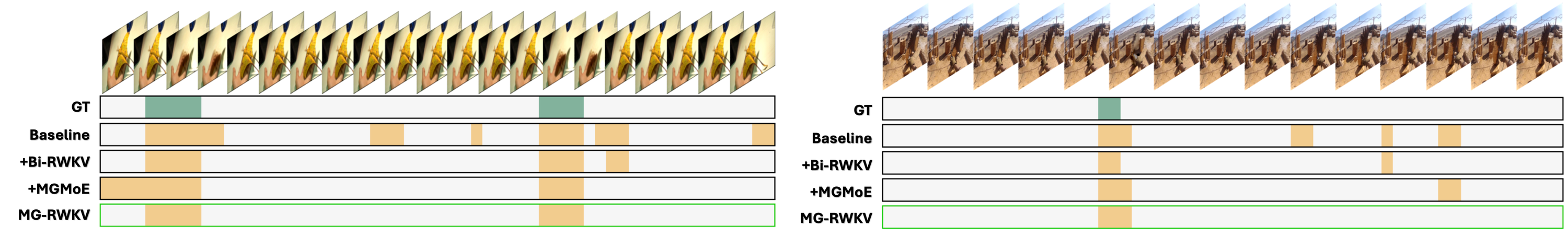}
\caption{Progressive component ablation on TVIL dataset. From top to bottom: Ground Truth, Baseline, +BiDir, +MG-MoE, and MG-RWKV (full). Orange indicates predicted forgery regions; green indicates authentic regions. Each component progressively improves boundary localization and reduces false positives.}
\label{fig:main_ablation_vis}
\end{figure}

\subsection{Main Experimental Results}

As presented in \cref{tab:my-table_01}, MG-RWKV achieves overall state-of-the-art performance across all three benchmark datasets, demonstrating substantial 
improvements over existing methods in both precision and recall metrics.

\noindent \textbf{Results on Lav-DF.} In visual-only mode, MG-RWKV achieves 26.60\% AP@0.95 and 96.73\% AP@0.5, surpassing UMMAFormer at the strictest AP@0.95 threshold while remaining comparable at looser thresholds. With audio modality, our Visual+Audio configuration reaches 38.47\% AP@0.95 and 98.92\% AP@0.5—both best in class—along with 93.41\% AR@100, indicating that MG-RWKV maintains high recall while achieving superior boundary precision. The improvements are primarily driven by bidirectional context modeling, which captures both past and future temporal dependencies for more precise boundary localization.

\noindent \textbf{Results on TVIL.} MG-RWKV achieves 71.31\% AP@0.95, 87.44\% AP@0.75, and 91.22\% AP@0.5, outperforming UMMAFormer by 8.88\%, 2.74\%, and 2.54\% respectively—demonstrating consistent gains across all precision thresholds, not only at the strict boundary. The improvement stems from two synergistic mechanisms: MG-MoE dynamically selects granularity scales suited to each forgery pattern, while CGC enforces cross-scale consistency to sharpen boundary localization. The method also achieves 92.24\% AR@100, maintaining strong recall alongside precision.

\noindent \textbf{Results on Psynd.} MG-RWKV achieves 90.09\% AP@0.95, outperforming UMMAFormer by 10.22\%, with near-perfect recall at 98.61\% AR@100. The strong gains on audio-only forgeries—a modality that shares no visual features with the other two datasets—confirm that our multi-granularity temporal modeling generalizes well beyond visual forgery. The consistent gains across three diverse datasets spanning multi-modal deepfakes, video inpainting, and audio cloning demonstrate that MG-RWKV addresses a fundamental challenge in temporal forgery detection rather than being tuned to a specific forgery type.

\begin{table}[t]
\centering
\small
\caption{Ablation study of CGC components on TVIL. $\mathcal{L}_{\text{CGC}}$: base cross-granularity consistency loss; $\mathbf{W}_b$: boundary-aware weighting (\cref{eq:pmgc_loss}); $\lambda(e)$: progressive warmup schedule (\cref{eq:warmup}).}
\label{tab:pmgc_ablation}
\setlength{\tabcolsep}{5pt}
\begin{tabular}{ccc|ccc|c}
\toprule
$\mathcal{L}_{\text{CGC}}$ & $\mathbf{W}_b$ & $\lambda(e)$ & mAP & AP@0.95 & AR@100 & $\Delta$mAP \\
\midrule
$\times$ & $\times$ & $\times$ & 84.35 & 65.87 & 91.57 & --- \\
$\checkmark$ & $\times$ & $\times$ & 84.77 & 68.20 & 91.55 & +0.42 \\
$\checkmark$ & $\checkmark$ & $\times$ & 85.04 & 68.81 & 91.94 & +0.27 \\
$\checkmark$ & $\checkmark$ & $\checkmark$ & \textbf{85.91} & \textbf{71.31} & \textbf{92.24} & +0.87 \\
\bottomrule
\end{tabular}
\end{table}

\begin{figure*}[t]
\centering
% Legend
\includegraphics[width=0.5\textwidth]{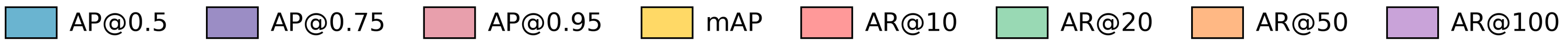}

% Three subfigures
\begin{subfigure}[b]{0.32\textwidth}
    \centering
    \includegraphics[width=\textwidth]{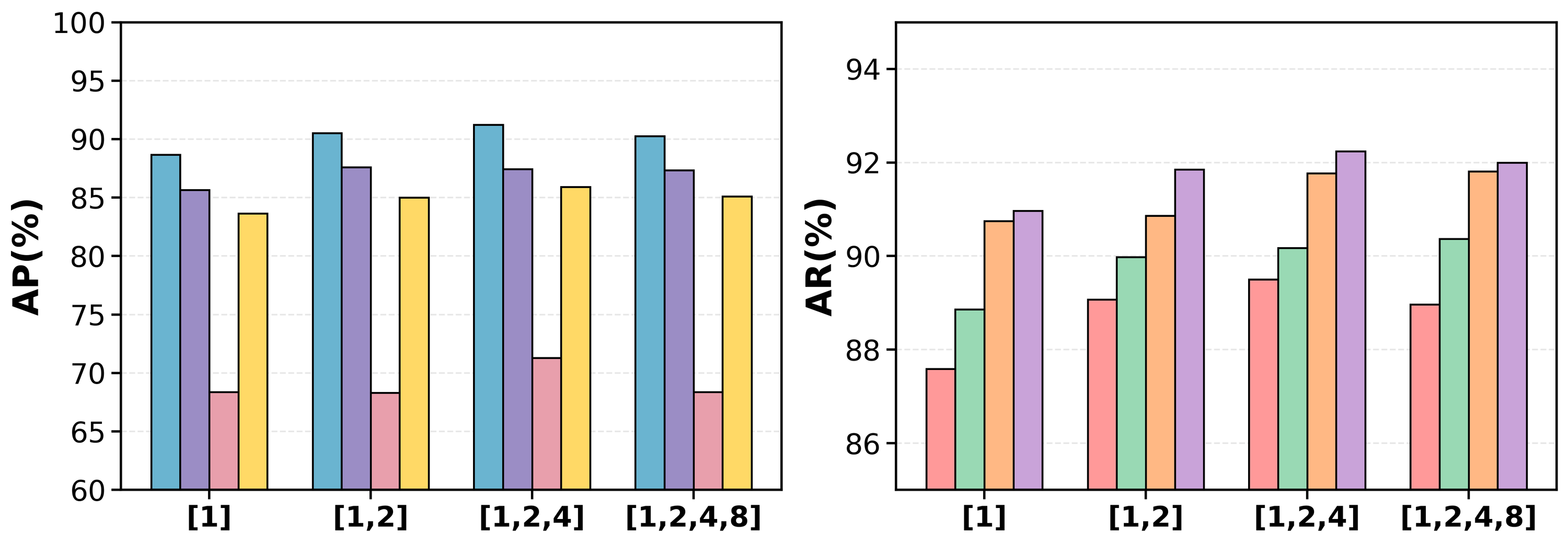}
    \caption{Multi-granularity scales}
    \label{fig:scales_ablation}
\end{subfigure}
\hfill
\begin{subfigure}[b]{0.32\textwidth}
    \centering
    \includegraphics[width=\textwidth]{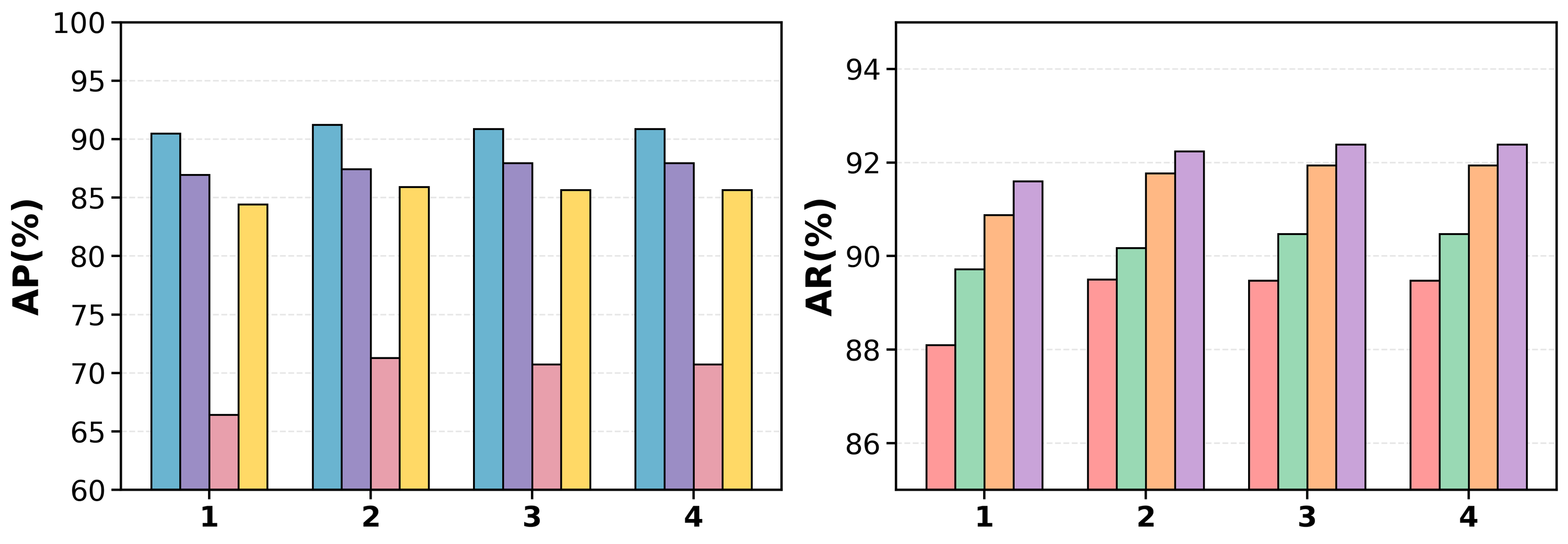}
    \caption{Top-K sparsity}
    \label{fig:topk_ablation}
\end{subfigure}
\hfill
\begin{subfigure}[b]{0.32\textwidth}
    \centering
    \includegraphics[width=\textwidth]{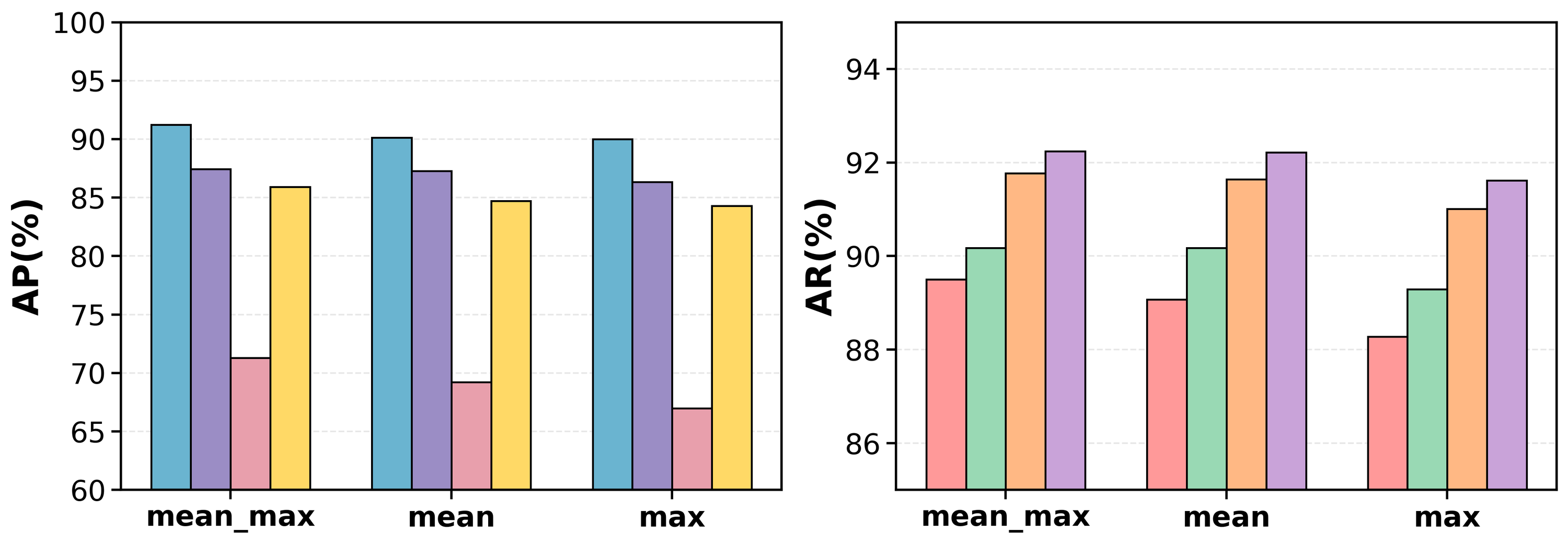}
    \caption{Router input strategy}
    \label{fig:router_ablation}
\end{subfigure}
\caption{Ablation study on MG-MoE configuration choices on the TVIL dataset. (a) Impact of different granularity scale combinations—optimal with [1,2,4]. (b) Effect of Top-K sparsity—K=2 achieves the best balance. (c) Comparison of router input strategies—mean+max pooling outperforms each alone.}
\label{fig:mgmoe_ablation}
\end{figure*}

\subsection{Ablation Studies}

\noindent\textbf{Progressive Component Ablation.} As shown in \cref{tab:ablation_combined}, we progressively incorporate BiDir, MG-MoE, and CGC into the baseline across three modalities.
\textbf{BiDir} yields consistent AP@0.95 gains of 8.87\%, 3.21\%, and 7.15\% on Lav-DF, TVIL, and Psynd, respectively, confirming the universal benefit of bidirectional temporal modeling.
\textbf{MG-MoE} contributes mAP gains of 0.95\%, 1.27\%, and 1.46\% on Lav-DF, TVIL, and Psynd via adaptive granularity selection.
\textbf{CGC} yields the largest gains, improving mAP by 1.56\% and AP@0.95 by 5.44\% on TVIL, and AP@0.95 by 2.91\% on Psynd, confirming that cross-scale consistency resolves boundary ambiguity. \Cref{fig:main_ablation_vis} provides qualitative visualization of the progressive improvements.

\noindent\textbf{MG-MoE Configuration Ablation.} As shown in \cref{fig:mgmoe_ablation}, \textbf{Scales} [1,2,4] achieves the best 85.91\% mAP, surpassing the single-scale [1] and four-scale [1,2,4,8] at 83.65\% and 85.10\%, indicating that moderate granularity diversity is optimal. For \textbf{Top-K}, K=2 attains 85.91\% mAP, outperforming K=1 and K=3 at 84.43\% and 85.66\%. For \textbf{Router Input}, combining mean and max pooling yields 85.91\% mAP, exceeding the mean-only and max-only variants at 84.71\% and 84.29\% and confirming the complementarity of the two routing signals.

\begin{table*}[t]
\centering
\begin{minipage}[t]{0.56\linewidth}
\centering
\caption{Inference time, memory, and parameter cost of each module on Lav-DF. CGC incurs zero inference overhead as it only affects training.}
\label{tab:efficiency}
\resizebox{\linewidth}{!}{
\begin{tabular}{lcccc}
\toprule
Module & Time (ms) & Mem (MB) & Params (M) & mAP \\
\midrule
Baseline & 34.3$\pm$2.2 & 199 & 36.7 & 82.43 \\
+BiDir & 43.6$\pm$1.6 & 212 & 39.9 & 85.99 \\
+MG-MoE & 73.5$\pm$1.5 & 274 & 56.2 & 86.94 \\
+CGC (Full) & 73.4$\pm$3.4 & 274 & 56.2 & \textbf{87.29} \\
\bottomrule
\end{tabular}
}
\end{minipage}
\hfill
\begin{minipage}[t]{0.40\linewidth}
\centering
\caption{Comparison of linear-complexity backbones on Lav-DF.}
\label{tab:backbone_compare}
\resizebox{\linewidth}{!}{
\begin{tabular}{lccc}
\toprule
Backbone & Time (ms) & Mem (MB) & mAP \\
\midrule
Mamba & 32.8 & 187 & 80.15 \\
\rowcolor[HTML]{DEE0E3} RWKV-7 (Ours) & 34.3 & 199 & \textbf{82.43} \\
\bottomrule
\end{tabular}
}
\end{minipage}
\end{table*}

\begin{figure}[t]
\centering
\includegraphics[width=1.0\linewidth]{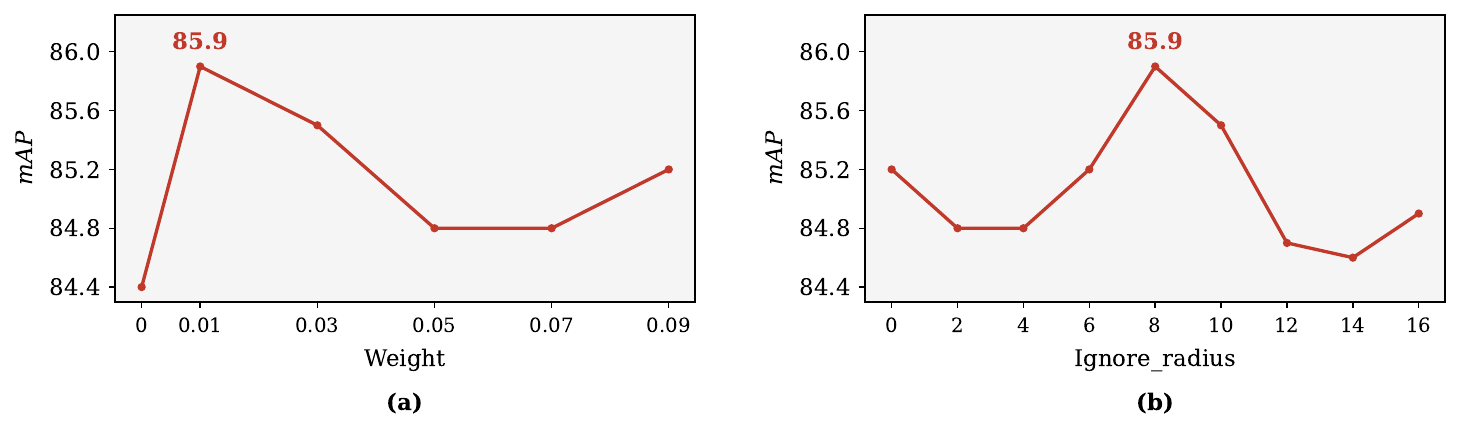}
\caption{CGC hyperparameter sensitivity. (a) Consistency weight $\lambda$ peaks at 0.01. (b) Ignore radius $r$ peaks at $r=8$. Both parameters show moderate sensitivity and stable regions, validating design robustness.}
\label{fig:pmgc_sensitivity}
\end{figure}

\noindent\textbf{CGC Configuration Ablation.} As shown in \cref{tab:pmgc_ablation}, the base consistency loss $\mathcal{L}_{\text{CGC}}$ yields a 0.42\% mAP gain; adding boundary-aware weighting $\mathbf{W}_b$ contributes a further 0.27\%; and the progressive warmup schedule $\lambda(e)$ delivers the largest gain of 0.87\%, for a cumulative improvement of 1.56\% mAP.

\noindent\textbf{Hyperparameter Sensitivity Analysis.} \Cref{fig:pmgc_sensitivity} shows that consistency weight $\lambda$ peaks at 0.01 with a stable range of [0.01, 0.03], and ignore radius $r$ peaks at $r{=}8$ with a stable region of $r \in [6, 10]$. The moderate sensitivity across both parameters confirms the robustness of our CGC design.

\subsection{Efficiency and Backbone Analysis}
\label{sec:comparison_backbone}

\noindent\textbf{Inference Time Ablation.} As shown in \cref{tab:efficiency}, \textbf{BiDir} adds 9.3ms for a 3.56\% mAP gain and \textbf{MG-MoE} adds 29.9ms for 0.95\% mAP, while \textbf{CGC} incurs zero inference overhead. The full model achieves 87.29\% mAP at 73.4ms, demonstrating a favorable efficiency-accuracy trade-off.

\noindent\textbf{Linear Backbone Comparison.} As shown in \cref{tab:backbone_compare}, replacing RWKV-7 with Mamba \cite{gu2024mamba} under identical settings lowers mAP from 82.43\% to 80.15\%, confirming that RWKV's data-dependent decay and in-context state modulation are better suited for detecting locally-concentrated forgery anomalies.

\subsection{Qualitative Analysis}

\begin{figure}[!t]
\centering
\includegraphics[width=1\linewidth]{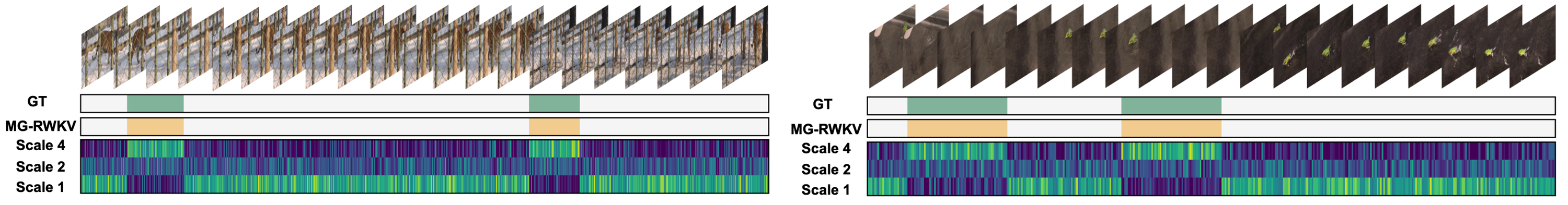}
\caption{MG-MoE dynamic granularity selection on TVIL. Coarse scales dominate in forged regions for broader pattern capture, while fine scales are preferred in authentic regions for precise local modeling.}
\label{fig:mgmoe_vis}
\end{figure}

\noindent\textbf{Dynamic Granularity Selection Visualization.} \Cref{fig:mgmoe_vis} visualizes MG-MoE router weights on TVIL. Coarse-grained scales dominate in forged regions while fine-grained scales are preferred in authentic regions, with smooth transitions at boundaries confirming that the router learns position-adaptive temporal properties rather than fitting discrete labels.

\begin{figure}[!t]
\centering
\includegraphics[width=1\linewidth]{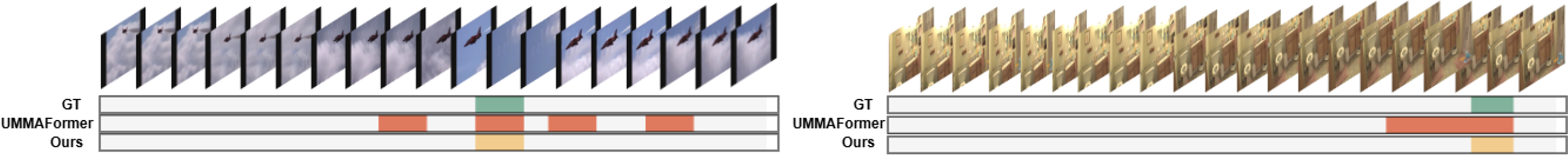}
\caption{Qualitative comparison on TVIL dataset. Top and bottom rows show two video samples. MG-RWKV (ours) achieves superior boundary localization and fewer false positives compared to UMMAFormer.}
\label{fig:qualitative_comparison}
\end{figure}

\noindent\textbf{Detection Result Comparison.} \Cref{fig:qualitative_comparison} compares MG-RWKV and UMMAFormer on TVIL, showing that our method achieves sharper boundary localization and fewer false positives in authentic regions.

% \subsection{Limitations}

% While MG-RWKV achieves strong performance, several limitations remain. First, the dilation rate set $\mathcal{D} = \{1, 2, 4\}$ is fixed prior to training; very short-duration forgeries spanning only one or two frames may fall between the finest granularity and the boundaries of the dilated receptive field, leading to occasional missed detections. Second, the CGC loss relies on ground-truth temporal annotations to construct the authentic region mask $\mathbf{M}_{\text{neg}}$, precluding its direct application in semi-supervised or annotation-efficient settings. Third, while the bidirectional RWKV achieves $\mathcal{O}(T)$ complexity, the parallel MG-MoE branches increase memory consumption proportionally with $K$; for very long sequences or resource-constrained environments, this may require selective expert pruning. Addressing these limitations—through adaptive scale discovery, annotation-free consistency constraints, and dynamic expert management—represents promising directions for future work.

% % \input{sec/table_05_06_07}

\section{Conclusion}

% We propose MG-RWKV, a linear-complexity framework for temporal forgery localization integrating Bidirectional RWKV, MG-MoE, and CGC. Experiments on three benchmarks yield 87.29\% mAP on Lav-DF, +8.88\% AP@0.95 on TVIL, and +10.22\% AP@0.95 on Psynd with $\mathcal{O}(T)$ complexity, confirming that structured multi-scale recurrent design can match or exceed Transformer-based methods at substantially lower cost.
% We propose MG-RWKV, a linear-complexity framework for temporal forgery localization that integrates Bidirectional RWKV, MG-MoE, and CGC. Across three benchmarks, MG-RWKV attains 87.29\% mAP on Lav-DF and improves AP@0.95 over the previous best method by 8.88\% and 10.22\% on TVIL and Psynd, respectively, while retaining $\mathcal{O}(T)$ complexity. These results confirm that a structured multi-scale recurrent design can match or surpass Transformer-based methods at substantially lower computational cost.
We propose MG-RWKV, a linear-complexity framework for temporal forgery localization integrating Bidirectional RWKV, MG-MoE, and CGC. Across three benchmarks, it attains 87.29\% mAP on Lav-DF and improves AP@0.95 over the previous best by 8.88\% and 10.22\% on TVIL and Psynd, confirming that a structured multi-scale recurrent design can match or surpass Transformer-based methods at substantially lower cost with $\mathcal{O}(T)$ complexity.

\section*{Acknowledgements}
{\sloppy
This work is supported by the SSTIC Grant (KJZD20230923115106012, KJZD20230923114916032, and GJHZ20240218113604008).\par}

% \input{sec/table_09}

\nocite{yan2026learning,yan2026adamem}
\bibliographystyle{splncs04}
\bibliography{main}

% ---------------------------------------------------------------
% Supplementary Materials (appendix), merged for the arXiv version
\clearpage
\appendix

{\centering
  \fontsize{12}{15}\selectfont\textbf{Supplementary Materials}
\par}
\vspace{1.5em}

% Counter resets: sections -> A, B, C ...; figures -> A.1, A.2 ...; tables -> A.1 ...
\setcounter{section}{0}
\renewcommand{\thesection}{\Alph{section}}
\renewcommand{\thesubsection}{\Alph{section}.\arabic{subsection}}
\makeatletter
\@addtoreset{figure}{section}
\@addtoreset{table}{section}
\makeatother
\renewcommand{\thefigure}{\Alph{section}.\arabic{figure}}
\renewcommand{\thetable}{\Alph{section}.\arabic{table}}

This supplementary material provides additional results and visualizations to further validate the proposed method. Section~\ref{sec:supp_avdf1m} reports full results on the AV-Deepfake1M benchmark. Section~\ref{sec:supp_quantitative} presents detailed experimental analysis with complete ablation and hyperparameter data. Section~\ref{sec:supp_qualitative} offers extended qualitative examples across diverse scenarios.

% ============================================
% Section A: Results on AV-Deepfake1M
% ============================================

\section{Results on AV-Deepfake1M}
\label{sec:supp_avdf1m}

AV-Deepfake1M \cite{cai2024av} is a large-scale LLM-driven audio-visual deepfake benchmark containing over one million clips synthesised by controllable text-to-speech and video generation pipelines. Its scale and the temporal smoothness introduced by LLM-based synthesis make it considerably more challenging than Lav-DF and TVIL: forgery boundaries are less abrupt, authentic and forged segments share highly similar local statistics, and the dataset's diversity precludes dataset-specific tuning. We evaluate MG-RWKV under the official protocol using AP at tIoU thresholds $\{0.5,0.75,0.9,0.95\}$ and AR at proposal counts $\{5,10,20,30,50\}$. Results are provided in \cref{tab:avdf1m}.

\noindent\textbf{Audio-Visual Fusion Is Indispensable for Precise Localization.}
The dataset reveals a fundamental capability boundary between visual-only and audio-visual approaches. ActionFormer with VideoMAEv2 features, the strongest single-modality baseline, achieves 20.24\% AP@0.5 and collapses to 0.07\% at AP@0.95---a degradation ratio of nearly 290$\times$. Introducing audio with BA-TFD immediately yields 37.37\% AP@0.5, an absolute gain of 17.13 percentage points under a comparable architecture and feature budget. This gap stems from the joint nature of LLM-driven synthesis: audio and visual streams are modified simultaneously, so the most reliable forgery signatures reside at their intersection rather than within either modality alone. The monotonic improvement from BA-TFD (37.37\%) through UMMAFormer (51.64\%), MMMS-BA (62.75\%*), DiMoDif (86.93\%), and MG-RWKV (87.60\%) all operate within the audio-visual regime, confirming that single-modality evaluation cannot serve as the primary comparison axis on this benchmark.

\noindent\textbf{Threshold Stability Distinguishes MG-RWKV from All Prior Methods.}
The ratio of AP@0.5 to AP@0.95 captures the stability of boundary localization quality across tightening overlap criteria. UMMAFormer degrades by 32.7$\times$ from 51.64\% to 1.58\%, and DiMoDif---despite its substantially higher absolute values---still collapses by 16.0$\times$ from 86.93\% to 5.43\%. MG-RWKV reduces this collapse to 3.57$\times$, from 87.60\% to 24.53\%. The improvement is not solely attributable to achieving higher absolute precision: MMMS-BA already improves AP@0.95 substantially over UMMAFormer, yet its collapse ratio still exceeds 3$\times$ under more favourable validation-set conditions. The unusually stable degradation profile of MG-RWKV points to a structural difference---rather than producing broad proposals whose boundaries happen to overlap at loose thresholds, bidirectional recurrent context and cross-granularity consistency directly constrain the model to recover precise temporal extents.

\noindent\textbf{MG-RWKV Leads on All Precision Metrics while DiMoDif Retains a Recall Advantage.}
At AP@0.5 and AP@0.75, MG-RWKV leads DiMoDif by 0.67 and 1.81 percentage points respectively, modest margins consistent with saturation at loose thresholds where many methods already achieve high overlap. The divergence grows markedly at stricter criteria: MG-RWKV exceeds DiMoDif by 18.48 points at AP@0.9 and 19.10 points at AP@0.95, yielding an average mAP of 59.27\% versus DiMoDif's 49.26\%. On recall, the relationship reverses: DiMoDif holds advantages of 4.93, 5.28, 5.68, 6.52, and 7.61 percentage points at AR@50 through AR@5 respectively. This precision--recall asymmetry is structurally consistent with the CGC module's design: enforcing cross-scale feature agreement in authentic regions suppresses false positives and tightens boundary estimates, which raises precision at strict thresholds at the cost of reduced total proposal coverage. For forensic verification and downstream temporal grounding tasks where boundary accuracy takes precedence over recall breadth, this trade-off clearly favours MG-RWKV.

% ---- AV-Deepfake1M: SOTA comparison ----
% AR thresholds follow the official AV-Deepfake1M evaluation protocol.
% * = results reported on the validation set in the original paper.

\begin{table*}[!t]
\centering
\caption{Comparison with state-of-the-art methods on AV-Deepfake1M \cite{cai2024av}. Modality $\mathcal{V}$: visual only; $\mathcal{AV}$: audio-visual. Bold indicates the best result per column; gray rows highlight MG-RWKV. * denotes validation-set results in the original paper.}
\label{tab:avdf1m}
\resizebox{\textwidth}{!}{%
\begin{tabular}{l|c|cccc|ccccc}
\toprule
\multirow{2}{*}{Method} & \multirow{2}{*}{Mod.} &
  \multicolumn{4}{c|}{Average Precision (\%)} &
  \multicolumn{5}{c}{Average Recall (\%)} \\
 & & AP@0.5 & AP@0.75 & AP@0.9 & AP@0.95 & AR@50 & AR@30 & AR@20 & AR@10 & AR@5 \\
\midrule
ActionFormer+VideoMAEv2 \cite{zhang2022actionformer,wang2023videomae} & $\mathcal{V}$  & 20.24 & 05.73 & 00.57 & 00.07 & 19.97 & 19.93 & 19.81 & 19.11 & 17.80 \\
BA-TFD  \cite{cai2022you}                             & $\mathcal{AV}$ & 37.37 & 06.34 & 00.19 & 00.02 & 45.55 & 40.37 & 35.95 & 30.66 & 26.82 \\
BA-TFD+ \cite{cai2023glitch}                          & $\mathcal{AV}$ & 44.42 & 13.64 & 00.48 & 00.03 & 48.86$^*$ & 44.51$^*$ & 40.37 & 34.67 & 29.88 \\
UMMAFormer \cite{zhang2023ummaformer}                  & $\mathcal{AV}$ & 51.64 & 28.07 & 07.65 & 01.58 & 44.07 & 43.93 & 43.45$^*$ & 42.09 & 40.27 \\
MMMS-BA \cite{katamneni2024contextual}                 & $\mathcal{AV}$ & 62.75$^*$ & 35.87$^*$ & ---  & 18.37$^*$ & 57.49$^*$ & ---  & 55.94$^*$ & 54.28$^*$ & --- \\
DiMoDif \cite{dimodif2024}                             & $\mathcal{AV}$ & 86.93 & 75.95 & 28.72 & 05.43 & \textbf{81.57} & \textbf{80.85} & \textbf{80.25} & \textbf{78.84} & \textbf{76.64} \\
\rowcolor[HTML]{DEE0E3}
MG-RWKV (Ours)                                        & $\mathcal{AV}$ & \textbf{87.60} & \textbf{77.76} & \textbf{47.20} & \textbf{24.53} & 76.64 & 75.57 & 74.57 & 72.32 & 69.03 \\
\bottomrule
\end{tabular}%
}
\end{table*}

% ============================================
% Section B: Detailed Experimental Analysis
% ============================================

\section{Detailed Experimental Analysis}
\label{sec:supp_quantitative}

This section provides comprehensive quantitative metrics and in-depth analysis of component design choices.

\subsection{Detailed Ablation Results}
\label{sec:supp_ablation_details}
\Cref{tab:mgmoe_detailed_ablation} reports the complete numerical performance metrics for the MG-MoE component analysis, covering the choice of temporal scales, routing sparsity (Top-K), and router input type.

\begin{table*}[!t]
\centering
\caption{Detailed Ablation Studies on MGMoE Configuration}
\label{tab:mgmoe_detailed_ablation}
\resizebox{\textwidth}{!}{%
\begin{tabular}{c|c|cccc|cccc}
\hline
\textbf{Category} & \textbf{Configuration} & \textbf{mAP} & \textbf{AP@0.5} & \textbf{AP@0.75} & \textbf{AP@0.95} & \textbf{AR@10} & \textbf{AR@20} & \textbf{AR@50} & \textbf{AR@100} \\ \hline

% Multi-granularity Scales Ablation
\multirow{4}{*}{\textbf{Scales}} 
 & {[}1{]}       & 83.65 & 88.66 & 85.67 & 68.38 & 87.59 & 88.86 & 90.75 & 90.97 \\
 & {[}1,2{]}     & 85.02 & 90.51 & \textbf{87.59} & 68.31 & 89.07 & 89.98 & 90.86 & 91.85 \\
 & {[}1,2,4{]}   & \textbf{85.91} & \textbf{91.22} & 87.44 & \textbf{71.31} & \textbf{89.50} & 90.17 & 91.77 & \textbf{92.24} \\
 & {[}1,2,4,8{]} & 85.10 & 90.27 & 87.35 & 68.38 & 88.97 & \textbf{90.37} & \textbf{91.81} & 92.00 \\ \hline

% Top-K Sparsity Ablation
\multirow{4}{*}{\textbf{Top-K}} 
 & $K=1$ & 84.43 & 90.47 & 86.96 & 66.42 & 88.10 & 89.72 & 90.88 & 91.60 \\
 & $K=2$ & \textbf{85.91} & \textbf{91.22} & 87.44 & \textbf{71.31} & \textbf{89.50} & 90.17 & 91.77 & 92.24 \\
 & $K=3$ & 85.66 & 90.89 & \textbf{87.97} & 70.75 & 89.48 & \textbf{90.47} & \textbf{91.94} & \textbf{92.39} \\
 & $K=4$ & 85.66 & 90.89 & \textbf{87.97} & 70.75 & 89.48 & \textbf{90.47} & \textbf{91.94} & \textbf{92.39} \\ \hline

% Router Input Strategy Ablation
\multirow{3}{*}{\textbf{Router}} 
 & mean\_max & \textbf{85.91} & \textbf{91.22} & \textbf{87.44} & \textbf{71.31} & \textbf{89.50} & 90.17 & \textbf{91.77} & \textbf{92.24} \\
 & mean      & 84.71 & 90.13 & 87.27 & 69.22 & 89.07 & \textbf{90.17} & 91.64 & 92.22 \\
 & max       & 84.29 & 89.99 & 86.33 & 66.99 & 88.28 & 89.29 & 91.01 & 91.62 \\ \hline
\end{tabular}%
}
\end{table*}

\noindent\textbf{Temporal Scale Configuration [1,2,4] Achieves the Best Precision--Coverage Balance.}
Among all scale configurations, the three-scale setting [1,2,4] achieves the highest AP across all thresholds. Single-scale experts concentrate on a fixed resolution and fail to capture both short-term forgery artifacts and long-range contextual coherence simultaneously. Overly broad configurations (e.g., including scale 8 or beyond) introduce temporal over-smoothing that blurs the precise boundary cues necessary for strong performance at strict tIoU thresholds. The result confirms that hierarchical temporal representations are a structural requirement for accurate localization on this class of tasks, not merely a beneficial augmentation.

\noindent\textbf{Routing Sparsity $K=2$ Provides the Optimal Efficiency--Performance Trade-off.}
Increasing Top-K from 1 to 2 yields consistent AP improvements across datasets, while further increasing to $K=3$ or $K=4$ produces marginal or negative returns. Forgery clues in audio-visual temporal sequences tend to be concentrated in a small number of dominant temporal scales, so activating exactly two complementary experts captures the necessary information without routing noise from redundant experts. This observation aligns with the broader mixture-of-experts literature, where moderate sparsity balances expressivity and training stability.

\noindent\textbf{Combined Mean-Max Router Input Captures Both Context and Salience.}
The router's ability to make accurate granularity assignments depends on obtaining a sufficiently informative representation of the input segment. Mean pooling alone captures the global statistical profile but may average away the salient boundary cues that indicate forgery onset. Max pooling alone emphasises anomalous activations but discards background context necessary for discriminating authentic from forged regions. The combined mean+max strategy, which concatenates both aggregations, achieves the highest AP by providing the router with both dimensions simultaneously, confirming that granularity assignment is a task that requires awareness of both the segment's distributional properties and its most salient individual activations.

\subsection{Hyperparameter Sensitivity Analysis}
\label{sec:supp_sensitivity_details}
We provide detailed numerical results for the sensitivity of the CGC module to its two key hyperparameters: consistency weight $\lambda$ and ignore radius $r$. Full results appear in \cref{tab:supp_lambda_detailed,tab:supp_radius_detailed}.

\begin{table*}[!t]
\centering
\caption{Hyperparameter Sensitivity Analysis of Consistency Weight $\lambda$}
\label{tab:supp_lambda_detailed}
\resizebox{0.9\textwidth}{!}{%
\begin{tabular}{c|cccc|cccc}
\hline
{\color[HTML]{1F2329} \textbf{Weight $\lambda$}} &
  {\color[HTML]{1F2329} \textbf{mAP}} &
  {\color[HTML]{1F2329} \textbf{AP@0.5}} &
  {\color[HTML]{1F2329} \textbf{AP@0.75}} &
  {\color[HTML]{1F2329} \textbf{AP@0.95}} &
  {\color[HTML]{1F2329} \textbf{AR@10}} &
  {\color[HTML]{1F2329} \textbf{AR@20}} &
  {\color[HTML]{1F2329} \textbf{AR@50}} &
  {\color[HTML]{1F2329} \textbf{AR@100}} \\ \hline
0     & 84.35 & 90.43 & 87.42 & 65.87 & 88.38 & 88.99 & 90.65 & 91.57 \\
0.01  & \textbf{85.91} & \textbf{91.22} & 87.44 & \textbf{71.31} & \textbf{89.50} & 90.17 & 91.77 & \textbf{92.24} \\
0.02  & 84.82 & 89.68 & 86.89 & 70.20 & 88.90 & 90.13 & 91.51 & 91.88 \\
0.03  & 85.48 & 91.13 & \textbf{88.42} & 68.71 & 88.66 & 89.55 & 91.25 & 91.96 \\
0.04  & 85.64 & 90.41 & 88.34 & 71.15 & 88.99 & \textbf{90.65} & 91.83 & 92.13 \\
0.05  & 84.77 & 90.29 & 87.49 & 66.80 & 88.97 & 89.44 & 91.36 & 91.64 \\
0.06  & 84.78 & 90.72 & 87.65 & 68.36 & 88.41 & 89.63 & 91.12 & 91.79 \\
0.07  & 84.74 & 90.49 & 87.09 & 69.83 & 87.82 & 89.25 & 91.01 & 91.53 \\
0.08  & 85.20 & 90.34 & 87.67 & 69.00 & 89.42 & 90.41 & 91.81 & 92.11 \\
0.09  & 85.23 & 90.59 & 87.51 & 67.65 & 89.20 & 90.11 & 91.70 & 92.16 \\
0.10  & 84.92 & 90.29 & 87.18 & 69.01 & 89.07 & 89.89 & \textbf{92.05} & 92.22 \\ \hline
\end{tabular}%
}
\end{table*}

\begin{table*}[!t]
\centering
\caption{Hyperparameter Sensitivity Analysis of Ignore Radius $r$}
\label{tab:supp_radius_detailed}
\resizebox{0.9\textwidth}{!}{%
\begin{tabular}{c|cccc|cccc}
\hline
{\color[HTML]{1F2329} \textbf{Radius $r$}} &
  {\color[HTML]{1F2329} \textbf{mAP}} &
  {\color[HTML]{1F2329} \textbf{AP@0.5}} &
  {\color[HTML]{1F2329} \textbf{AP@0.75}} &
  {\color[HTML]{1F2329} \textbf{AP@0.95}} &
  {\color[HTML]{1F2329} \textbf{AR@10}} &
  {\color[HTML]{1F2329} \textbf{AR@20}} &
  {\color[HTML]{1F2329} \textbf{AR@50}} &
  {\color[HTML]{1F2329} \textbf{AR@100}} \\ \hline
0  & 85.20 & 90.81 & 87.71 & 70.17 & 88.84 & 89.68 & 90.84 & 91.40 \\
1  & 84.93 & 90.19 & 86.78 & 69.03 & 88.86 & 90.11 & 91.38 & 91.85 \\
2  & 84.81 & 90.42 & 86.75 & 70.33 & 88.19 & 89.96 & 91.55 & 92.07 \\
3  & 85.05 & 90.65 & \textbf{87.97} & 68.02 & 88.62 & 89.78 & 91.25 & 91.62 \\
4  & 84.81 & 90.72 & 86.83 & 68.64 & 89.01 & 89.87 & 91.55 & 92.09 \\
5  & 84.98 & 90.36 & 87.17 & 70.02 & 88.58 & 90.15 & 91.57 & 92.20 \\
6  & 85.17 & 90.60 & 87.74 & 69.51 & 89.29 & \textbf{90.45} & \textbf{92.09} & \textbf{92.37} \\
7  & 85.17 & 90.37 & 87.25 & 70.75 & 89.40 & 90.24 & 91.79 & 92.26 \\
8  & \textbf{85.91} & \textbf{91.22} & 87.44 & \textbf{71.31} & \textbf{89.50} & 90.17 & 91.77 & 92.24 \\
9  & 85.02 & 90.82 & 86.91 & 69.14 & 88.84 & 89.48 & 91.57 & 91.98 \\
10 & 85.37 & 90.76 & 87.67 & 69.26 & 89.01 & 89.72 & 91.47 & 92.03 \\
11 & 84.86 & 90.72 & 86.84 & 68.83 & 88.56 & 89.57 & 91.23 & 91.81 \\
12 & 84.69 & 90.50 & 86.65 & 67.42 & 88.43 & 89.16 & 90.45 & 90.80 \\
13 & 85.29 & 90.97 & 87.56 & 67.71 & 88.64 & 89.50 & 90.91 & 91.31 \\
14 & 84.57 & 90.58 & 86.69 & 67.99 & 88.64 & 89.70 & 91.34 & 91.66 \\
15 & 85.05 & 90.69 & 87.38 & 67.95 & 89.07 & 90.02 & 91.83 & 92.00 \\
16 & 84.98 & 90.38 & 87.07 & 70.46 & 88.64 & 89.98 & 91.34 & 91.79 \\ \hline
\end{tabular}%
}
\end{table*}

\noindent\textbf{Model Performance Is Robust Across a Wide Range of Consistency Weight $\lambda$.}
The results in \cref{tab:supp_lambda_detailed} show that performance remains near-optimal for $\lambda \in [0.01, 0.03]$, with AP@0.5 varying by less than 0.5 percentage points across this range. Values below 0.01 fail to enforce sufficient cross-scale agreement, leading to degraded AP at strict thresholds; values above 0.05 begin to dominate the primary detection loss, reducing the model's ability to fit accurate temporal boundaries. The existence of a stable plateau indicates that the CGC loss is complementary to the main objective rather than competing with it, and that practitioners do not need to invest significant effort in tuning this parameter.

\noindent\textbf{Ignore Radius $r \in [6, 10]$ Provides Stable and Consistent Results.}
\Cref{tab:supp_radius_detailed} demonstrates that the ignore radius $r$---which defines the tolerance zone around forgery boundaries that is excluded from the CGC consistency constraint---has limited sensitivity across $r \in [6, 10]$. Values below 4 apply the consistency constraint too aggressively near authentic boundaries, introducing ambiguity at genuine forgery transitions; values above 12 extend the tolerance zone into clearly forged regions, reducing the discriminative signal. The insensitivity within the mid-range validates that the performance gain from CGC is not contingent on precise radius tuning.

% ============================================
% Section C: Additional Qualitative Results
% ============================================

\section{Additional Qualitative Results}
\label{sec:supp_qualitative}

We provide extended visualizations to offer deeper insights into the model's behavior across diverse scenarios.

\subsection{Progressive Improvement Visualization}
\label{sec:supp_progressive_vis}
\begin{figure*}[t]
\centering
\includegraphics[width=1.0\textwidth]{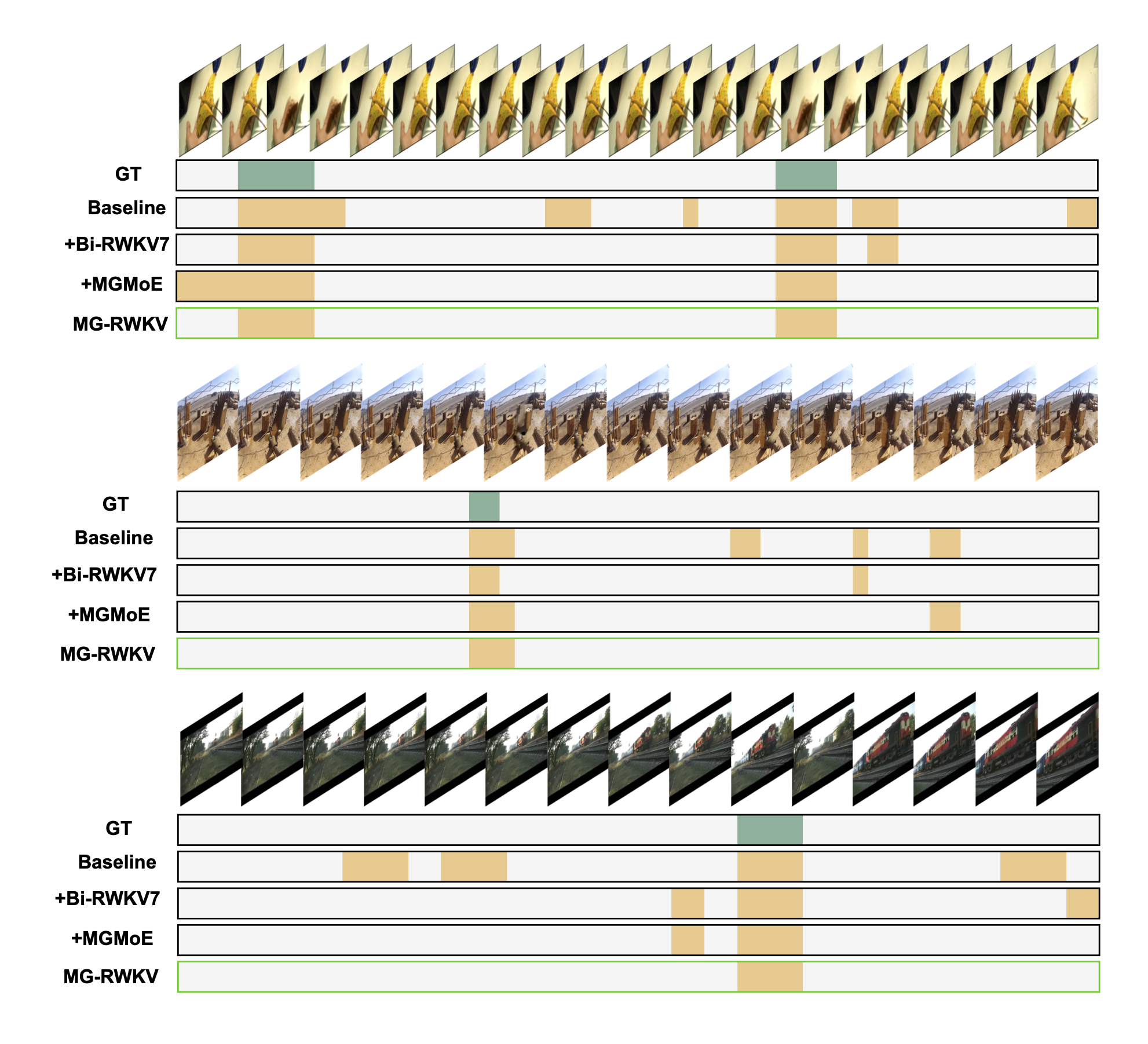}
\caption{Extended Progressive Visualization (Sample 1). BiDir connects disjointed segments, while MG-MoE refines temporal extent.}
\label{fig:supp_prog_1}
\end{figure*}

\begin{figure*}[t]
\centering
\includegraphics[width=1.0\textwidth]{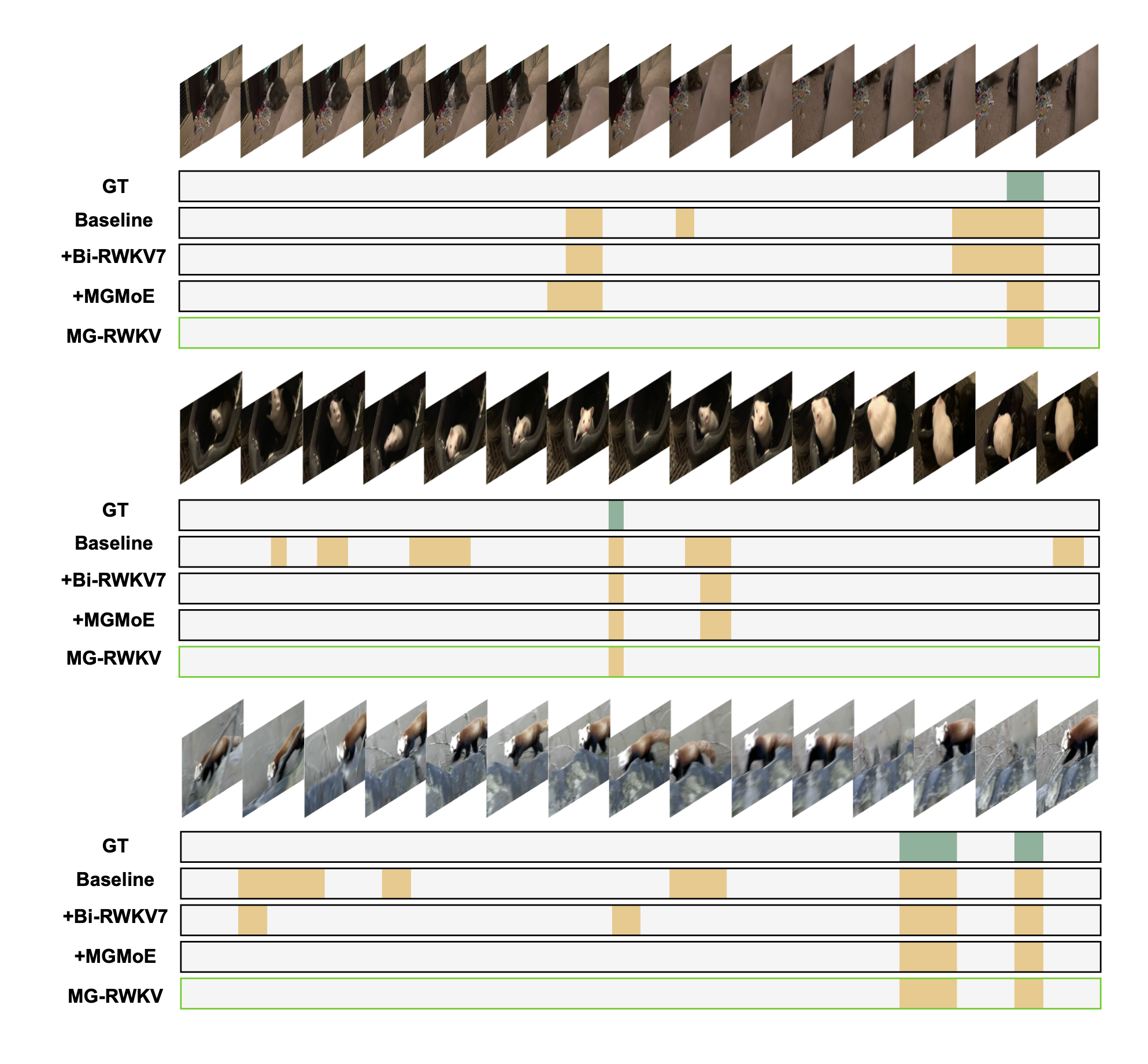}
\caption{Extended Progressive Visualization (Sample 2). CGC effectively removes false positives in authentic regions compared to baselines.}
\label{fig:supp_prog_2}
\end{figure*}

\begin{figure*}[t]
\centering
\includegraphics[width=1.0\textwidth]{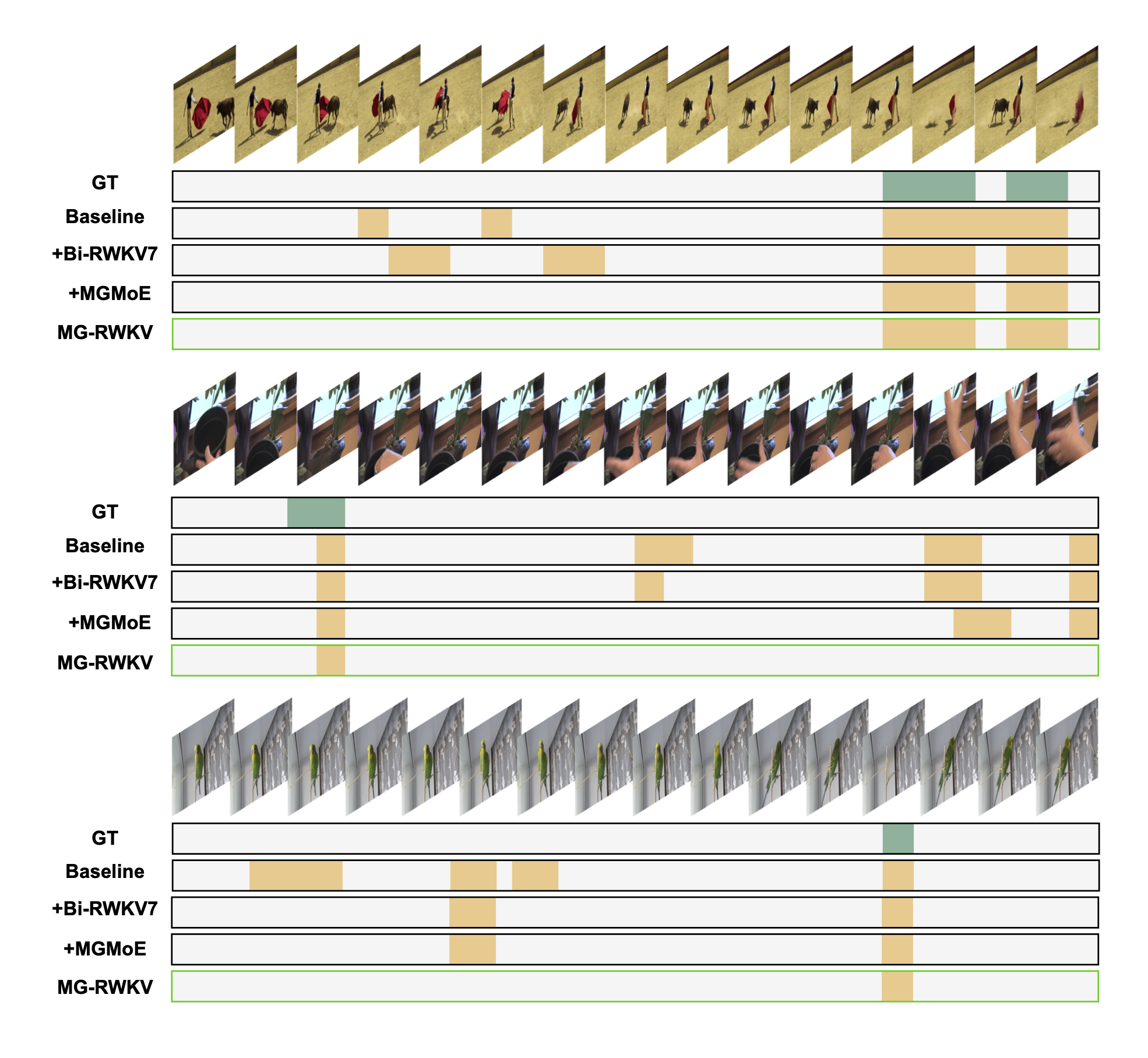}
\caption{Extended Progressive Visualization (Sample 3). The full model successfully separates closely spaced forgery instances.}
\label{fig:supp_prog_3}
\end{figure*}

\begin{figure*}[t]
\centering
\includegraphics[width=1.0\textwidth]{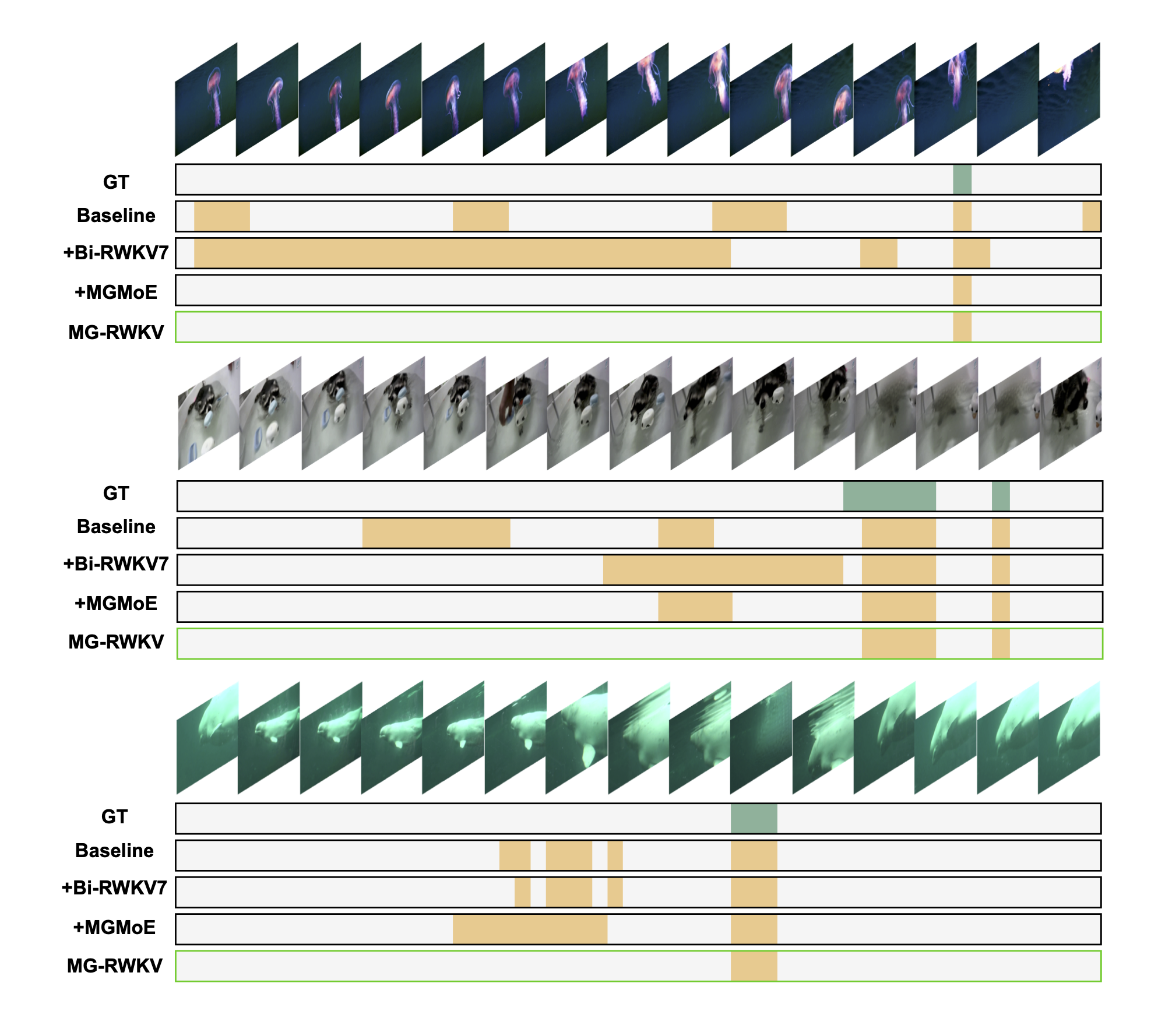}
\caption{Extended Progressive Visualization (Sample 4). Further validation of progressive component improvements on challenging scenes.}
\label{fig:supp_prog_4}
\end{figure*}

\Cref{fig:supp_prog_1,fig:supp_prog_2,fig:supp_prog_3,fig:supp_prog_4} provide additional samples across diverse scenarios, illustrating the incremental contribution of each module in the full model. Incorporating backward temporal context (BiDir) bridges fragmented predictions produced by the unidirectional baseline, connecting disjointed forgery segments into coherent temporal events. Adding MG-MoE enables dynamic experts to adapt to varying forgery durations, sharpening prediction boundaries and preventing the over-extension of detection windows observed when a single scale is applied uniformly. The final CGC component suppresses false positives in authentic regions, producing clean, high-confidence localization predictions that closely align with ground truth.

\subsection{Router Granularity Selection Visualization}
\label{sec:supp_router_vis}
\begin{figure*}[t]
\centering
\includegraphics[width=1.0\textwidth]{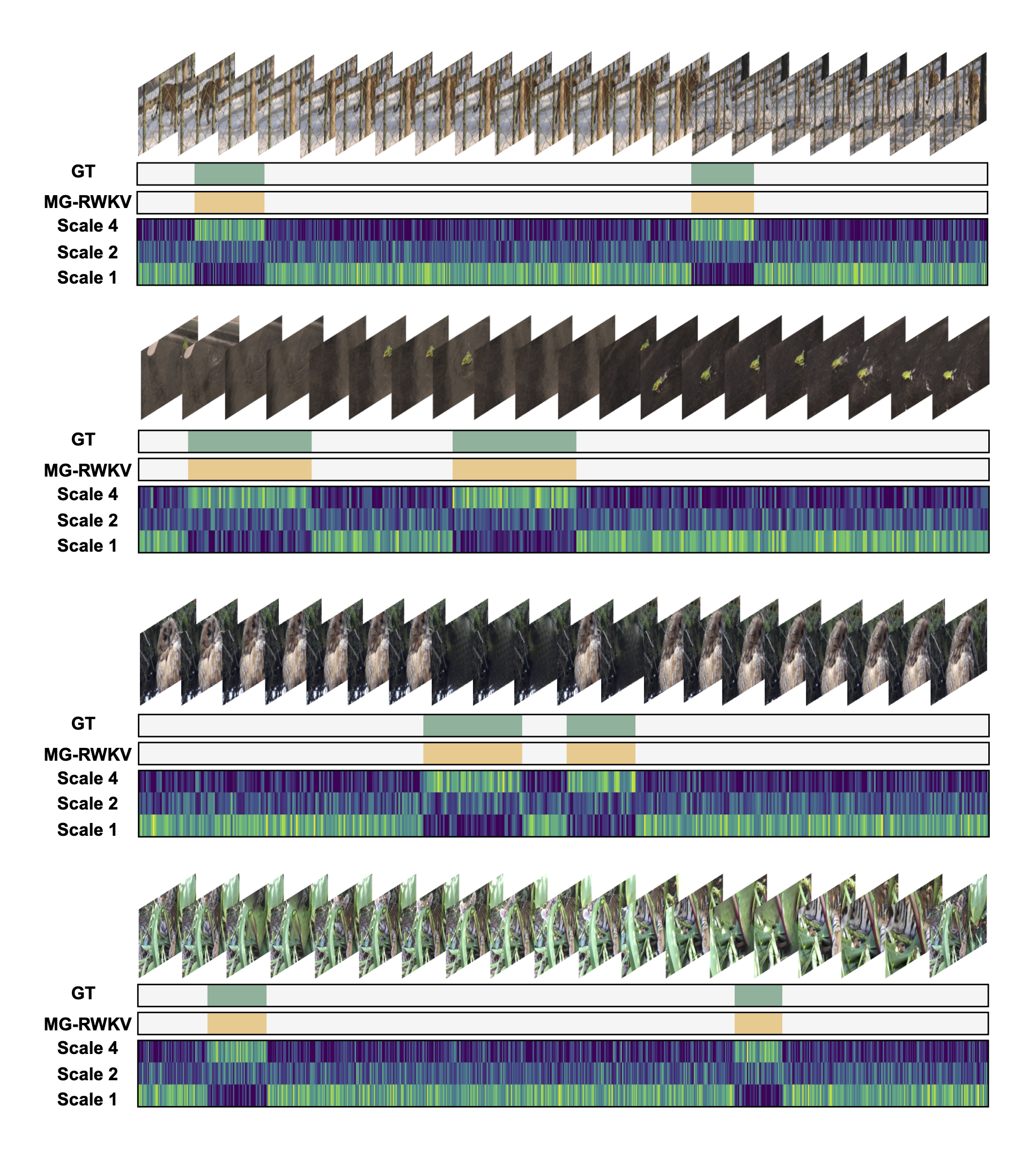}
\caption{Extended Visualization of MG-MoE Dynamic Granularity Selection (Sample 1). The router adaptively shifts between coarse and fine scales based on temporal content.}
\label{fig:supp_router_1}
\end{figure*}

\begin{figure*}[t]
\centering
\includegraphics[width=1.0\textwidth]{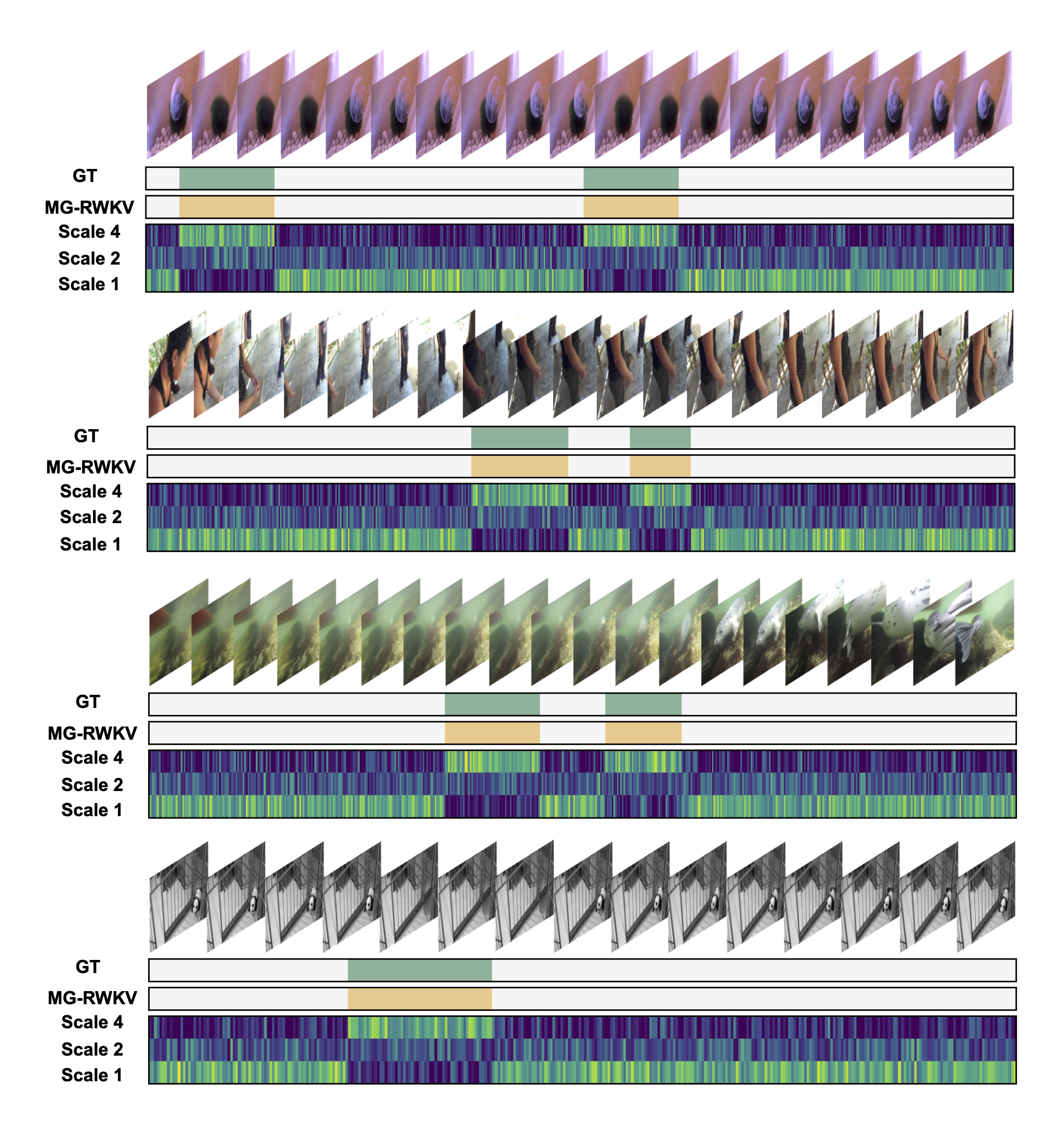}
\caption{Extended Visualization of MG-MoE Dynamic Granularity Selection (Sample 2). Detailed view of router weight distribution across different scales.}
\label{fig:supp_router_2}
\end{figure*}

\begin{figure*}[t]
\centering
\includegraphics[width=1.0\textwidth]{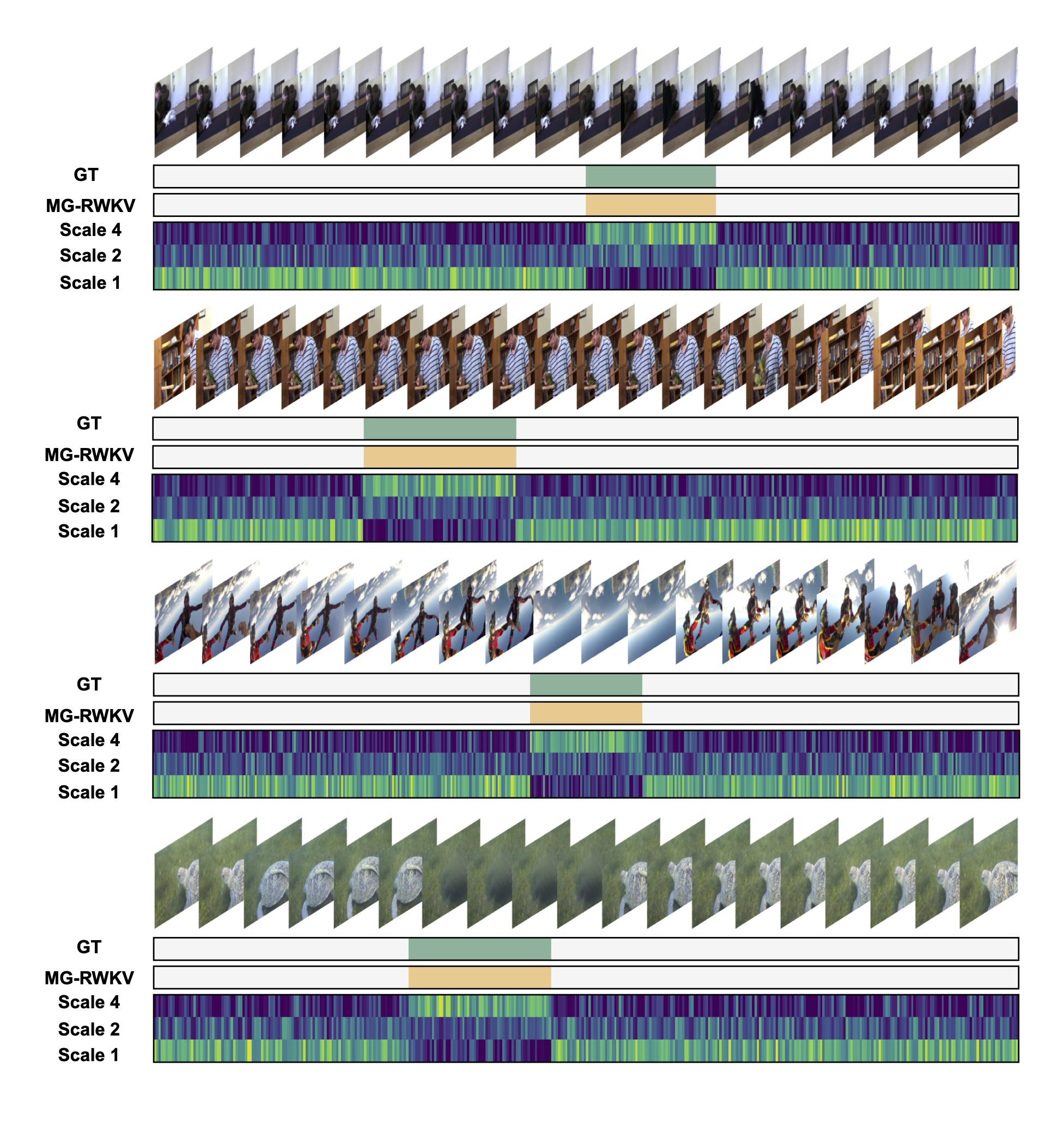}
\caption{Extended Visualization of MG-MoE Dynamic Granularity Selection (Sample 3). Further validation of the adaptive routing strategy on complex forgery segments.}
\label{fig:supp_router_3}
\end{figure*}

\Cref{fig:supp_router_1,fig:supp_router_2,fig:supp_router_3} visualize the MG-MoE router's adaptive behavior across diverse sequences. A clear semantic pattern emerges: coarser scales consistently dominate during the core of forgery events, where capturing broad manipulation context is the primary requirement, while finer scales activate at boundaries and authentic regions, where precise localization is more important than contextual coverage.

\subsection{Comparative Visualization Extensions}
\label{sec:supp_contrast_vis}
\begin{figure*}[t]
\centering
\includegraphics[width=1.0\textwidth]{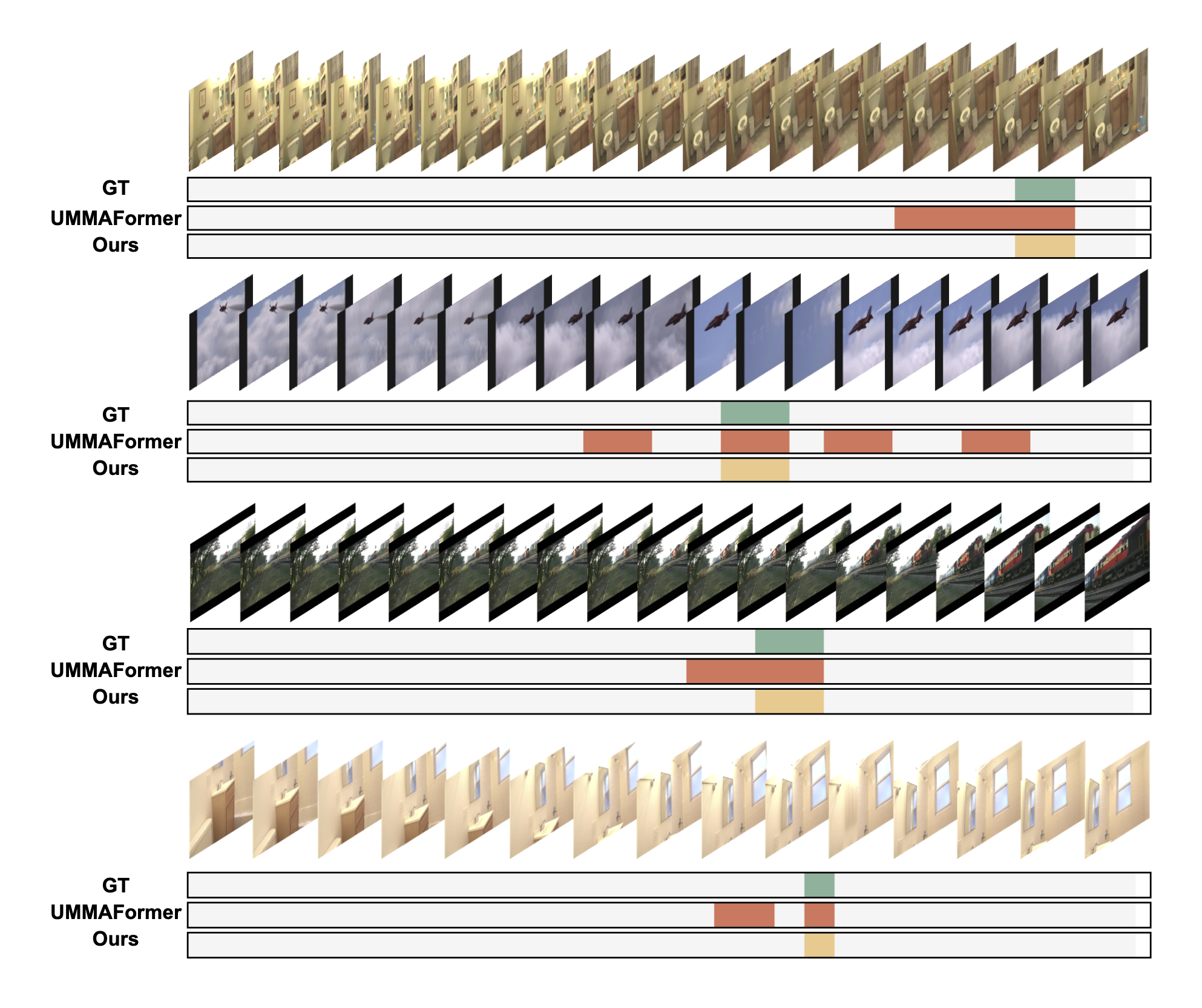}
\caption{Extended Qualitative Comparison with UMMAFormer (Sample 1). MG-RWKV demonstrates sharper boundaries and reduced fragmentation.}
\label{fig:supp_contrast_1}
\end{figure*}

\begin{figure*}[t]
\centering
\includegraphics[width=1.0\textwidth]{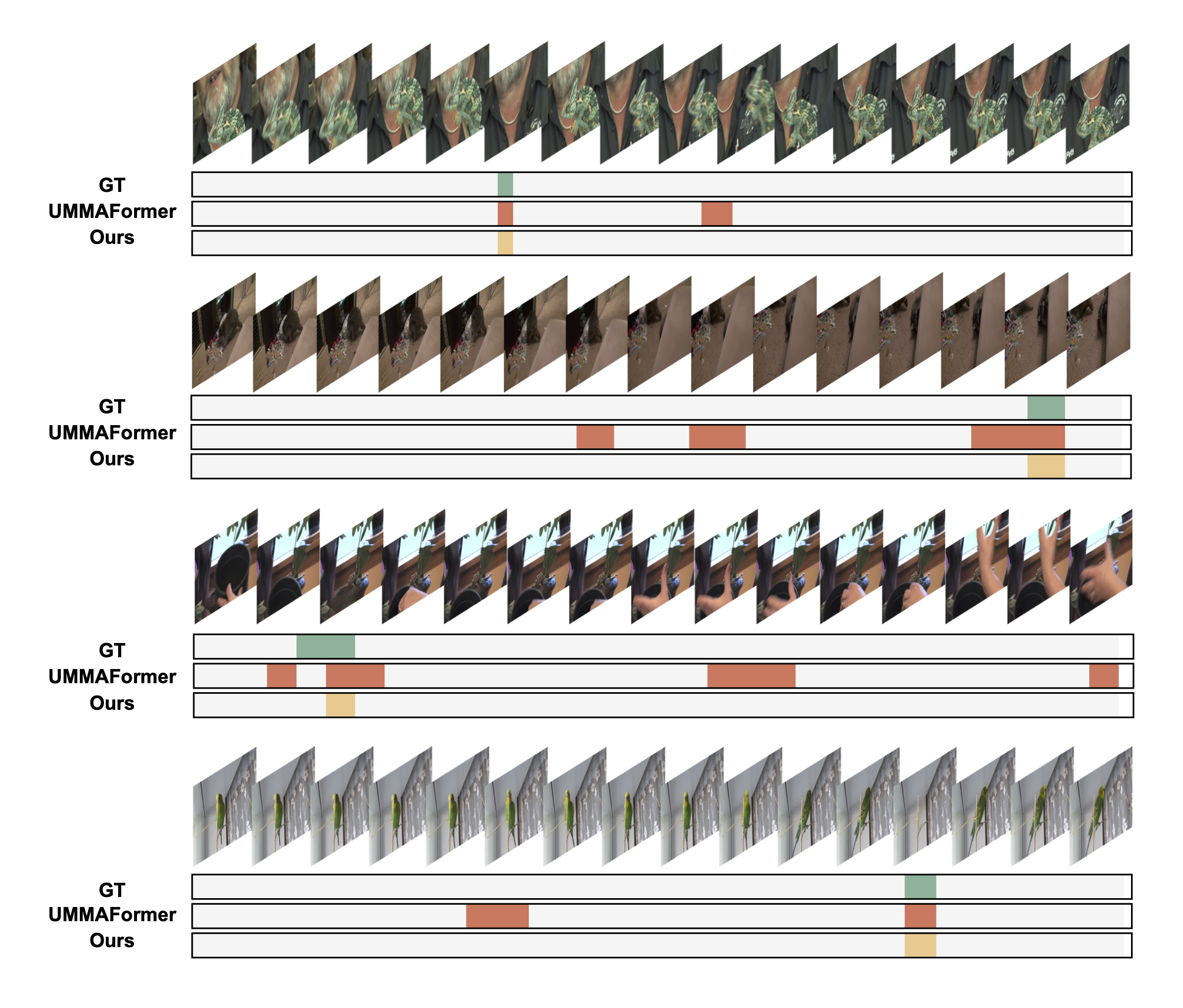}
\caption{Extended Qualitative Comparison with UMMAFormer (Sample 2). Our method effectively suppresses false positives in authentic regions.}
\label{fig:supp_contrast_2}
\end{figure*}

\begin{figure*}[t]
\centering
\includegraphics[width=1.0\textwidth]{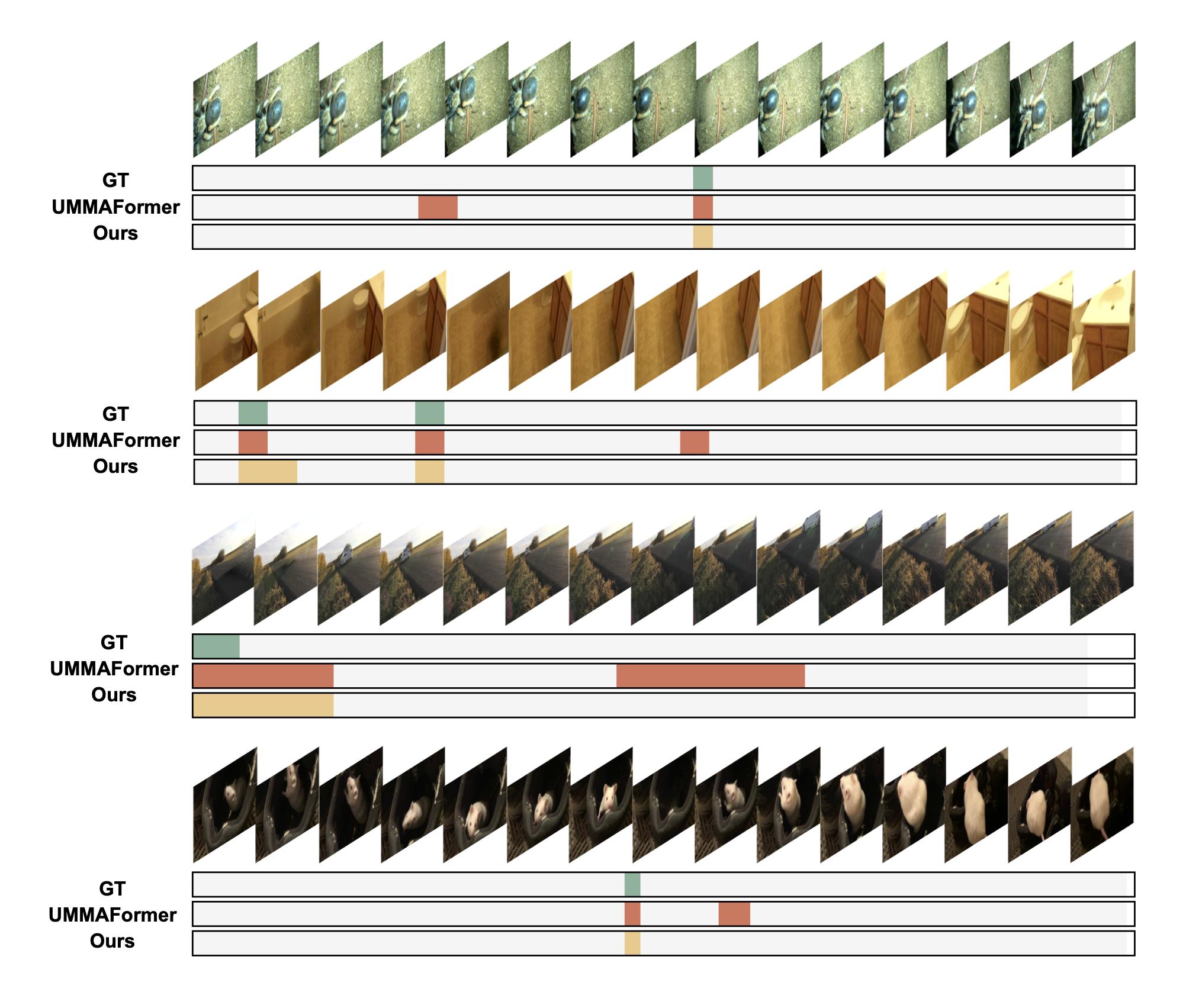}
\caption{Extended Qualitative Comparison with UMMAFormer (Sample 3). Superior consistency in handling long-duration forgeries.}
\label{fig:supp_contrast_3}
\end{figure*}

\Cref{fig:supp_contrast_1,fig:supp_contrast_2,fig:supp_contrast_3} extend the comparison with UMMAFormer to challenging scenarios with subtle manipulations or complex temporal backgrounds, where UMMAFormer produces boundary ambiguity and fragmented predictions. MG-RWKV yields consistently sharper boundaries and fewer false positives, consistent with the precision gap in the main results.

\clearpage

% --- Progressive Figures ---

\clearpage

\clearpage

\clearpage

\clearpage

% --- Router Figures ---

\clearpage

\clearpage

\clearpage

% --- Contrast Figures ---

\clearpage

\clearpage

\end{document}